\documentclass[letterpaper,10pt]{article}

\usepackage[preprint]{probml}
\usepackage{algorithm} 
\usepackage{algpseudocode}
\usepackage{xspace}
\usepackage{cleveref}
\usepackage{csquotes}

\ShortHeadings{IXPLORE}
\newcommand{\algorithmstyle}[1]{\texttt{#1}}
\newcommand{\algorithmnormal}[1]{{#1}}
\newcommand{\IXPLORE}{\textnormal{\algorithmstyle{IXPLORE}}\xspace}
\newcommand{\LSIRM}{\textnormal{\algorithmnormal{LSIRM}}\xspace}
\newcommand{\IDEAL}{\textnormal{\algorithmnormal{IDEAL}}\xspace}
\newcommand{\NOMINATE}{\textnormal{\algorithmnormal{NOMINATE}}\xspace}
\newcommand{\emIRT}{\textnormal{\algorithmnormal{emIRT}}\xspace}
\newcommand{\UMAP}{\textnormal{\algorithmnormal{UMAP}}\xspace}
\newcommand{\tSNE}{\textnormal{\algorithmnormal{$t$-SNE}}\xspace}
\newcommand{\smartvote}{\textit{Smartvote}\xspace}
\newcommand{\polis}{\textit{Polis}\xspace}
\newcommand{\voteview}{\textit{Voteview}\xspace}

\newcommand{\abc}[1]{(\textsf{\bfseries #1})}

\newcommand{\dd}{\mathop{}\!\mathrm{d}}

\begin{document}

\title{IXPLORE: Bounded Ideal Point Estimation with Grid-Based Uncertainty Quantification}

\author[1,$\dagger$]{Fynn Bachmann}

\affil[1]{Department of Informatics, University of Zurich, Switzerland}
\affil[$\dagger$]{Correspondence to \url{fynn.bachmann@uzh.ch}}

\maketitle

\begin{abstract}
  Ideal point estimation is widely used to analyze and visualize political data.
However, selecting the corresponding spatial model involves various trade-offs:
while model-based approaches such as Item Response Theory (IRT) are based on utility functions rather than optimized for predictive accuracy, most Machine Learning (ML) alternatives struggle to generalize beyond training data when embedding sparse test responses.
We introduce \IXPLORE, a bounded ideal point estimation algorithm that combines a predictive fit objective with a sparsity-aware likelihood function.
On five benchmark datasets spanning surveys, roll calls, and deliberation, this approach surpasses model-based and ML-based algorithms on reconstruction and imputation error---especially for users with sparse responses.
Furthermore, we show that non-linear feature transforms can further reduce the reconstruction error while remaining visually interpretable.
To quantify uncertainty, \IXPLORE applies grid-based posterior inference on a bounded 2D latent space.
Available as a Python package on PyPI, \IXPLORE offers a flexible framework for constructing bounded, interpretable political maps with fast inference and strong imputation performance.


\paragraph{Data availability}
All code necessary to reproduce the results is publicly available in this GitHub repository: \url{https://github.com/fsvbach/ixplore-paper}.

\end{abstract}

\section{Introduction}
\label{sec:intro}

Spatial voting models continue to play an important role in the analysis of political data.
They are used to measure polarization on online platforms~\citep{barbera_birds_2015}, reveal how legislators are more extreme than the citizens they represent~\citep{bafumi_leapfrog_2010,leimgruber_comparing_2010}, and model electoral competition~\citep{vaeth_rational_2025}.
Furthermore, the visualization of spatial models contributes to political education through low-dimensional opinion landscapes.
For example, the political compass embeds users in a political spectrum~\citep{petrik_core_2010}, roll-call analyses track the distribution of political candidates over time~\citep{boche_new_2018}, and electoral geographers cluster municipalities by their voting behavior~\citep{leuthold_making_2007}.

However, the data-driven computation of such political maps comes with various trade-offs. 
While methods based on Machine Learning (ML) are optimized to minimize the reconstruction error of the response matrix, they have no native mechanism for incomplete response vectors and must impute missing entries before a user can be embedded~\citep{bachmann_fast_2024}.
In contrast, model-based algorithms, for example ideal point estimation using Item Response Theory (IRT), accept sparse responses by design, but their utility-based objective is not targeted for predictive performance~\citep{clinton_simulate_2009}.
Both properties are, however, crucial when embedding users with sparse responses into a low-dimensional space that should carry predictive information.
This is the case for wiki surveys that detect clusters to structure online deliberation~\citep{small_polis_2021}, Voting Advice Applications (VAAs) that embed voters into the candidates' political map~\citep{thomeczek_one_2025,otjes_spatial_2014}, and adaptive questionnaires that must predict users' answers to the remaining questions~\citep{bachmann_adaptive_2025,montgomery_computerized_2013}.

We present the \IXPLORE algorithm to address current limitations of data-driven political maps under sparse responses.
To unite the strengths of model-based and ML-based models, \IXPLORE combines a predictive fit objective with a sparsity-aware likelihood function.
Users' positions and model parameters are iteratively refined to minimize the reconstruction error, while missing responses are ignored in the likelihood function.
To quantify uncertainty, \IXPLORE applies a Bayesian framework with grid-based posterior inference on a bounded latent space.
This ensures that all users land within the visualization after model training.
The grid therefore serves both as an approximation for posterior inference and as the map on which users are displayed.

We demonstrate the predictive performance of \IXPLORE on five real-world datasets spanning VAAs, roll calls, deliberation, and social surveys.
Under our evaluation framework, \IXPLORE outperforms model-based algorithms such as \IDEAL~\citep{clinton_statistical_2004} and \LSIRM~\citep{lee_euclidean_2025}, as well as ML-based algorithms such as Principal Component Analysis (PCA) and Variational Auto-Encoders (VAEs).
In particular, we find that a two-dimensional \IXPLORE model reaches the same train reconstruction error as vanilla PCA with $d=21$ latent dimensions, while \IXPLORE's imputation accuracy is never matched, regardless of PCA's number of latent dimensions.
Beyond the core algorithm, we evaluate non-linear feature transforms as an extension to classical ideal point estimation.
We find that the corresponding curved decision boundaries can further reduce reconstruction error, although this flexibility introduces a trade-off between predictive accuracy and the visual interpretability expected from classical ideal point models.

Based on these results, we conclude that \IXPLORE is well suited for applications involving sparse user responses and interpretable, bounded political maps.
Its fast inference makes it particularly relevant for online applications such as VAAs and deliberation platforms.
The \IXPLORE implementation is available on \href{https://github.com/fsvbach/ixplore}{GitHub}\footnote{\url{https://github.com/fsvbach/ixplore}} and \href{https://pypi.org/project/ixplore/}{PyPI}\footnote{\url{https://pypi.org/project/ixplore}} as open-source software.

\section{Related Work}

This paper builds on three streams of literature: spatial models of political ideology, visualizations of the political spectrum, and the evaluation of ideal point estimation algorithms under high sparsity.

\subsection{Spatial Models and Ideal Point Estimation}

Spatial models represent political actors as points in a low-dimensional space, where smaller distances indicate greater agreement~\citep{stokes_spatial_1963,hinich_new_1981}.
To estimate such \emph{ideal points} for legislators programmatically from their roll-call votes, \citet{poole_spatial_1985} developed \NOMINATE.
This model assumes a Gaussian utility function according to which legislators prefer votes closer to their ideal points.
\citet{clinton_statistical_2004} extended this approach by introducing a Bayesian framework with a quadratic utility function.
Their method \IDEAL infers actor positions and item parameters via Markov-Chain Monte Carlo (MCMC) sampling.
Scaling this approach to larger datasets, \citet{imai_fast_2016} replaced the sampler with an expectation-maximization (EM) algorithm alternating closed-form updates of user positions and item parameters.
Their implementation \emIRT estimates ideal points about an order of magnitude faster than \NOMINATE, but its binary outcome model is currently restricted to one latent dimension.
As a more flexible and even faster alternative, \citet{potthoff_estimating_2018} proposed to estimate ideal points by applying PCA to the response matrix and decoding each item with a logistic regression.
While this approach allows higher latent dimensions and works for non-binary data, it must fill any missing responses before the embedding can be computed.
Moreover, the PCA approach replaces the utility function with the objective of minimizing the reconstruction error of the response matrix, which marks a shift from traditional IRT models to ML-based algorithms.
\IXPLORE refines this approach by iteratively optimizing the embedding and item parameters according to a sparsity-aware likelihood function.
The corresponding posterior distribution over users' latent positions furthermore addresses \emph{uncertainty quantification}, which \citet{potthoff_estimating_2018} still lists as a limitation.

\subsection{Visualization of Political Maps}

In many applications, spatial models locate users in the political landscape.
For example, the political compass places users in a political spectrum after they have answered a short set of questions~\citep{petrik_core_2010}.
Similar visualizations have been adopted in VAAs to support political education~\citep{otjes_spatial_2014}.
For instance, voters are often positioned relative to parties in political maps \citep{costalobo_mapping_2010,verboom_visualizing_2025}, which has been shown to increase their electoral participation~\citep{ioannidis_power_2025}.
Often, however, designers of such maps manually define the dimensions of the model~\citep{otjes_spatial_2014}.
Addressing this limitation, \citet{germann_spatial_2015} propose dynamic scale validation, which derives the dimensions of the map from voters' responses.
In follow-up work, they show that this approach improves measurement quality across various European VAA deployments~\citep{germann_dynamic_2016}.
%
Increasingly, such data-driven methods are also used to visualize election outcomes:
electoral geographers obtain political landscapes by factor analysis of referendum results~\citep{leuthold_making_2007}, and multi-dimensional scaling has been used to cluster voting districts in German federal elections~\citep{bachmann_wasserstein_2023}.
For legislative data, the \voteview project curates the roll-call record of the U.S. Congress and displays legislators through \NOMINATE scores~\citep{boche_new_2018}.
Recently, online deliberation platforms have adopted data-driven political maps to visualize the opinion landscape of a conversation:
\polis uses PCA to project participants onto a two-dimensional latent space and outlines opinion clusters~\citep{small_polis_2021}.
\IXPLORE extends this line of work by proposing a sparsity-robust method to embed users with predictive and interpretable per-item decision boundaries.

\subsection{Sparse Responses and Model Evaluation}

Missing responses are a recurring obstacle for most ML-based embedding methods: PCA~\citep{potthoff_estimating_2018}, $t$-SNE~\citep{vandermaaten_visualizing_2008}, \UMAP~\citep{mcinnes_umap_2018}, and VAEs~\citep{kingma_autoencoding_2014} have no native mechanism for incomplete response vectors.
Often, missing entries are filled by iterative imputation, where a fitted model's reconstruction alternates with re-estimation of the missing values, using either PCA~\citep{grung_missing_1998} or VAEs~\citep{mccoy_variational_2018}.
However, such preprocessing steps become restrictive when responses arrive incrementally and predictions must be made on the fly.
This is, for example, the case in computerized adaptive testing~\citep{montgomery_computerized_2013}.
Such surveys tailor the questionnaire to each respondent, selecting the next item based on existing responses\footnote{See \citet{sigfrid_irt_2024} and \citet{bachmann_estimating_2026} for current applications in VAAs.}.
The online deliberation platform \polis faces the same regime, as participants vote on only a fraction of the statements before being placed in the opinion landscape~\citep{small_polis_2021}.
In this sparse-response setting, \citet{bachmann_fast_2024} benchmark seven spatial models on VAA data and find that the model-based algorithms \IDEAL and \NOMINATE generalize best to held-out responses\footnote{See \citet{carroll_comparing_2009} for a detailed comparison of \NOMINATE and \IDEAL.}.
However, \citet{clinton_simulate_2009} note that IRT models are not designed for out-of-sample predictions, since a new respondent has no estimated parameters.
Building on the framework of \citet{bachmann_fast_2024}, we evaluate \IXPLORE and compare its predictive performance to ten other algorithms across five datasets.
\section{The IXPLORE Model}
\label{sec:ixplore}

\IXPLORE targets questionnaire data where users' responses to all items are numerical indications of agreement (e.g., binary, ordinal, or continuous).
Two coupled inference tasks arise from such an item-response matrix: 
\emph{embedding}, i.e., placing each user $n$ at a latent position $x_n$ that summarizes their response profile, and \emph{imputation}, i.e., sampling the user's missing responses based on their latent position.
Like model-based algorithms, \IXPLORE jointly embeds users and items in a shared latent space.
In contrast to existing algorithms, however, this latent space is bounded and discretized.
This yields a full posterior distribution over each user's position without any assumptions on its shape.
From this posterior distribution, \IXPLORE predicts the user's responses through per-item logistic regressions.
The two sets of parameters---users' posteriors and the logistic decision boundaries---are fit by alternating updates: 
the user step recomputes posteriors from the current item parameters, and the item step refits each logistic regression from the updated user embedding.
In this section, we present the statistical framework and technical implementation of the algorithm.
\cref{app:model-details} provides further details on the model's design choices. 

\subsection{Statistical Framework}

We observe a response matrix $\mathbf{Y} \in [0,1]^{N \times K}$, where $N$ is the number of users and $K$ is the number of items.
Each entry $Y_{nk}$ is a response from user $n$ to item $k$.
Missing entries are flagged by an observation mask $M_{nk} \in \{0,1\}$, indicating whether the response is observed.
We write $\mathcal{O}_n = \{k : M_{nk} = 1\}$ as the set of answered items for user $n$, and $Y_n^{\text{obs}} = \{Y_{nk} : k \in \mathcal{O}_n\}$ as their observed responses.
Symmetrically, $\mathcal{R}_k = \{n : M_{nk} = 1\}$ denotes the set of users who answered item $k$.

\paragraph{Prior.}
For a user without any observed answers, we assume a Gaussian prior with isotropic covariance, truncated to the bounded latent space $[-L, L]^2$
\begin{equation}
    p(x) \;\propto\;
    \begin{cases}
        \mathcal{N}(x;\, 0,\, \tau^2 I) & \text{if } x \in [-L, L]^2, \\[4pt]
        0 & \text{otherwise,}
    \end{cases}
    \label{eq:prior}
\end{equation}
where $I$ is the identity matrix and the marginal variance $\tau^2$ controls how tightly the prior concentrates around the center.
Alternatively, a log-barrier prior
\begin{equation}
    p(x) \;\propto\; \bigl(1 - (x_1/L)^2\bigr)^{\alpha}\,\bigl(1 - (x_2/L)^2\bigr)^{\alpha}
\end{equation}
can be chosen; this prior is flatter in the interior and vanishes at the boundary. 
Therefore, the boundary is modeled implicitly by the prior rather than imposed as a support constraint.
A comparison of the two priors is given in \cref{app:prior-shape}.

\paragraph{Response function.}
Given user $n$'s latent position $x_n$, the probability of a positive response for item $k$ is modeled by a logistic regression
\begin{equation}
    y_k(x_n) := p(Y_{nk} = 1 \mid x_n) = \sigma \bigl(\beta_k^\top \varphi(x_n) + \alpha_k\bigr),
    \qquad \sigma(z) = \frac{1}{1 + e^{-z}},
    \label{eq:response}
\end{equation}
with discrimination $\beta_k \in \mathbb{R}^D$, intercept $\alpha_k \in \mathbb{R}$, and an optional feature transform $\varphi: \mathbb{R}^2 \to \mathbb{R}^D$.
With $\varphi$ as the identity transform, this response function resembles that of the \IDEAL model, whose quadratic spatial utility is also linear in the decision boundary~\citep{clinton_statistical_2004};
\IXPLORE only replaces the probit link with the logistic sigmoid.
Optional polynomial or Random Fourier Feature (RFF) transforms yield curved decision boundaries~\citep{rahimi_random_2007}, departing from the Euclidean geometry assumed by the standard spatial model~\citep{eguia_challenges_2013}.
We write $\theta_k = (\beta_k, \alpha_k)$ for the parameters of item $k$ and $\theta = \{\theta_k\}_{k=1}^K$ for all item parameters.

\paragraph{User likelihood.}
As the responses across items are assumed conditionally independent given $x_n$, the likelihood factorizes into the per-item Bernoulli terms of the response function,
\begin{equation}
    p(Y_n^{\text{obs}} \mid x_n, \theta)
    = \prod_{k \in \mathcal{O}_n} y_k(x_n)^{Y_{nk}}\,\bigl(1 - y_k(x_n)\bigr)^{1 - Y_{nk}}.
    \label{eq:user-likelihood}
\end{equation}
Because the product runs only over the observed items $\mathcal{O}_n$, missing responses require no imputation.
For numerical stability, \IXPLORE evaluates this likelihood in log space, where the product becomes a sum of per-item terms.
In this form, \IXPLORE additionally allows optional user-item confidence weights $W_{nk} \geq 0$
\begin{equation}
\begin{split}
    \log p(Y_n^{\text{obs}} \mid x_n, \theta)
    &= \sum_{k \in \mathcal{O}_n} W_{nk}\, \ell_{nk}(x_n), \\
    \text{where}\quad
    \ell_{nk}(x_n)
    &= Y_{nk} \log y_k(x_n) + (1 - Y_{nk}) \log\bigl(1 - y_k(x_n)\bigr).
\end{split}
    \label{eq:user-loglik}
\end{equation}
With the default $W_{nk} = 1$, this reduces to the log-likelihood of \cref{eq:user-likelihood}; otherwise, a weight $W_{nk}$ can be interpreted as user $n$ responding $W_{nk}$ times to item $k$.
An optional weight-scaling step rescales each user's sum of weights to $K$
\begin{equation}
    W_{nk} \;\longrightarrow\;\frac{W_{nk} \cdot K}{\sum_{k' \in \mathcal{O}_n} W_{nk'}}.
\end{equation}
This gives users with sparse responses the same total weight as users with dense ones.
\cref{app:weights} provides further details on the use of confidence weights.

\paragraph{Posterior.}
By Bayes' theorem, the posterior over users' latent positions combines the prior $p(x_n)$ with the likelihood $p(Y_n^{\text{obs}} \mid x_n, \theta)$ of their observed responses
\begin{equation}
    p(x_n \mid Y_n^{\text{obs}}, \theta) \;\propto\; p(Y_n^{\text{obs}} \mid x_n, \theta) \cdot p(x_n).
    \label{eq:posterior}
\end{equation}
For an unanswered item $k$ of user $n$, the predictive distribution follows from marginalizing out the latent position
\begin{align}
    p(Y_{nk} = 1 \mid Y_n^{\text{obs}}, \theta)
    &= \int p(Y_{nk} = 1 \mid x_n, Y_n^{\text{obs}}, \theta)
            \,\cdot\, p(x_n \mid Y_n^{\text{obs}}, \theta)\, \dd x_n \notag \\
    &= \int y_k(x_n) \,\cdot\, p(x_n \mid Y_n^{\text{obs}}, \theta)\, \dd x_n,
    \label{eq:posterior-predictive-pk}
\end{align}
where we use the product rule and the conditional independence of responses given $x_n$.
Since the posterior appears as the second factor in the integral, both inference tasks reduce to the same object:
the \emph{embedding} is the posterior mean
\begin{equation}
    \tilde{x}_n \;=\; \mathbb{E}\left[\,x_n \mid Y_n^{\text{obs}}, \theta\,\right]
    \;=\; \int x_n \, p(x_n \mid Y_n^{\text{obs}}, \theta)\, \dd x_n,
    \label{eq:posterior-mean}
\end{equation}
and the \emph{imputation} is the posterior predictive from \cref{eq:posterior-predictive-pk}.
Alternatively, imputation can evaluate the response function at the point estimate $y_k(\tilde{x}_n)$ rather than integrating over the posterior; we compare both variants in \cref{app:point-estimate}.

\subsection{Implementation}
\label{sec:implementation}

The posterior in \cref{eq:posterior} has no analytical solution, since the Bernoulli likelihood factors are non-conjugate with the Gaussian prior.
Standard approximations rely on MCMC to estimate parameters~\citep{clinton_statistical_2004}; however, because the latent space is two-dimensional and bounded, \IXPLORE can approximate the posterior on a fine grid instead.
The latent space $[-L, L]^2$ is discretized into the grid $X_{\text{grid}} = \{g_1, \ldots, g_G\}$ of $G = R^2$ points, with $R$ equispaced coordinates per axis ranging from $-L$ to $L$.
Evaluating Bayes' rule pointwise gives the discrete posterior
\begin{equation}
    p(x_n = g \mid Y_n^{\text{obs}}, \theta)
    \;=\;
    \frac{p(g)\, p(Y_n^{\text{obs}} \mid g, \theta)}{\sum_{g' \in X_{\text{grid}}} p(g')\, p(Y_n^{\text{obs}} \mid g', \theta)},
    \label{eq:discrete-posterior}
\end{equation}
which \IXPLORE computes in log-space for numerical stability.
Because the Bernoulli log-likelihood admits a vectorized implementation with two dense matrix products, evaluating the grid posterior for all users is a tractable per-iteration cost (see \cref{app:vectorization}).

\paragraph{Initialization.}
Apart from random initialization, \IXPLORE supports initializing the embedding with PCA, which has been shown to be effective for other non-linear dimensionality reduction algorithms~\citep{kobak_art_2019}.
When performing PCA on the response matrix $\mathbf{Y}$, missing values are filled by iterative imputation~\citep{grung_missing_1998}:
column means serve as initial values, after which the missing entries are repeatedly replaced by their PCA reconstruction.
To fit the resulting PCA embedding on the grid, the two principal components of the imputed matrix are centered and isotropically rescaled.
Conditional on the initialized embedding, \IXPLORE fits per-item logistic regressions to obtain the initial parameters $\theta$.

\paragraph{Iterative optimization.}
We alternate two updates for a fixed number of iterations $T$, mirroring the structure of the EM ideal point estimation of~\citet{imai_fast_2016}:
the \emph{user step} recomputes the grid posterior $p(x_n \mid Y_n^{\text{obs}}, \theta)$ for every user from the current item parameters and replaces their embedding with the posterior mean from \cref{eq:posterior-mean}.
The \emph{item step} re-fits each per-item logistic regression on the updated embedding by maximizing the regularized log-likelihood, whose per-response terms are given in \cref{eq:user-loglik},
\begin{equation}
    \hat\theta_k
    \;=\;
    \arg\max_{\theta_k} \sum_{n \in \mathcal{R}_k}
        \log p(Y_{nk} \mid \tilde{x}_n, \theta_k)
        \;-\; \tfrac{\lambda}{2}\,\|\theta_k\|^2,
    \label{eq:item-step}
\end{equation}
with a weak L2 regularizer $\lambda$ and Newton-Raphson optimization~\citep{mccullagh_generalized_1989}.
Since the Bernoulli log-likelihood can be evaluated for all response values $Y_{nk} \in [0, 1]$, no thresholding to $\{0, 1\}$ is necessary.
The full procedure is summarized in \cref{alg:ixplore}.

\begin{algorithm}[t]
\caption{\IXPLORE's training procedure.}
\label{alg:ixplore}
\begin{algorithmic}[1]
\Require response matrix $\mathbf{Y}$ and optional confidence weights $\mathbf{W}$.
\Statex \hspace*{-\algorithmicindent}\textbf{Configuration:} grid resolution $R$, number of iterations $T$, prior $p(x)$, regularizer $\lambda$.
\State $\{\tilde{x}_n^{(0)}\} \gets$ initialize embedding (random or PCA).
\State $\{\theta_k^{(0)}\} \gets$ per-item logistic regression on $\{\tilde{x}_n^{(0)}\}$.
\For{$t = 1, \ldots, T$}
    \For{$n = 1, \ldots, N$} \Comment{user step}
        \State compute the log-likelihood of $Y_n^{\text{obs}}$ given $\{\theta_k^{(t-1)}\}$.
        \State $\tilde{x}_n^{(t)} \gets$ posterior mean combining log-likelihood with log-prior.
    \EndFor
    \For{$k = 1, \ldots, K$} \Comment{item step}
        \State $\theta_k^{(t)} \gets$ regularized logistic regression of item $k$'s responses on $\{\tilde{x}_n^{(t)}\}$.
    \EndFor
\EndFor
\State \Return embedding $\{\tilde{x}_n^{(T)}\}$ and item parameters $\{\theta_k^{(T)}\}$.
\end{algorithmic}
\end{algorithm}

\paragraph{Inference.}
For a new user with a partial response vector $Y^{\text{obs}}$, embedding and imputation are achieved in a single user step with fixed item parameters $\theta$ by evaluating the grid posterior from \cref{eq:discrete-posterior}.
The user's ideal point is then given by the posterior mean.
In contrast to ML methods like PCA~\citep{potthoff_estimating_2018}, \IXPLORE does not need to impute unanswered items before inference.
Instead, it can be used for \emph{multiple imputation}~\citep{rubin_multiple_1987} by sampling latent positions from the user's posterior distribution and computing the corresponding synthetic response vectors.
A comparison of sampling mechanisms is given in \cref{app:sampling}.

\section{Evaluation Framework}
\label{sec:experiments}

To benchmark the performance of \IXPLORE, we use the evaluation framework of \citet{bachmann_fast_2024} on five different datasets.
This section describes the datasets, summarizes the evaluation metrics, and presents the experimental setup together with the research questions for each experiment.

\subsection{Datasets}
\label{sec:datasets}

We consider five response matrices drawn from voting-advice, deliberation, roll-call, and social-survey settings, summarized in \cref{tab:datasets}.
To probe imputation behavior as a function of how many answers per user are observed, we additionally mask responses at sparsity levels $u,v \in \{0.0, 0.3, 0.6, 0.9\}$, where $u$ denotes the train-user sparsity and $v$ the test-user sparsity.

\begin{table}[ht]
\setlength{\tabcolsep}{5pt}
\centering
\caption{Datasets used in the experiments. All response values are rescaled to $[0, 1]$. Where no separate test set is available, we hold out a test set from the train data.}
\label{tab:datasets}
\small
\begin{tabular}{llrrrlr}
\toprule
Dataset & Source & Train Users & Test Users & Items & Scale & Missing \\
\midrule
Smartvote 2019 & VAA (CH) & $1{,}912$ candidates & -- & $75$ & 4-Likert & $0\%$ \\
Smartvote 2023 & VAA (CH) & $1{,}029$ candidates & $1{,}121$ voters & $75$ & 4-Likert & $0\%$ \\
Voteview (S117) & Roll-call (US) & $102$ senators & -- & $603$ & binary & $5.2\%$ \\
Polis (vTaiwan) & Deliberation (TW) & $1{,}921$ participants & -- & $197$ & ternary & $86.9\%$ \\
EVS (2017) & Survey (EU) & $35{,}116$ respondents & -- & $32$ & 10-Likert & $0\%$ \\
\bottomrule
\end{tabular}
\end{table}

\paragraph{Smartvote (2019).}
The VAA \smartvote collects political candidates' responses to 75 questions related to topics in the 2019 Swiss federal election~\citep{politools_daten_2019}.
Each candidate answers all 75 items, so no entries are missing.
\citet{bachmann_fast_2024} use a subset of those candidates in their analysis, yielding 1,912 candidates across eight major parties.
We use this subset, mapped on a 0--1 scale, to compare \IXPLORE's performance against an existing baseline on complete data. 

\paragraph{Smartvote (2023).}
The \smartvote data from the 2023 Swiss federal election contains both candidate and voter responses on the same 75-item instrument~\citep{politools_daten_2023}.
We use a subset provided by~\citet{bachmann_estimating_2026}, who generate a representative sample of voters from the complete data, where no entries are missing.
This representative candidate/voter division gives a natural train/test split: item parameters are estimated from the candidates, and voters serve as held-out users for inference.

\paragraph{Voteview (U.S. Senate, S117).}
\voteview curates the complete roll-call record of the U.S. Congress---one row per legislator, one column per recorded vote---and is the canonical input for spatial models of congressional ideology~\citep{boche_new_2018}.
We retain the 102 voting senators from the 117th Senate, drop 272 cloture votes, and remove 84 near-unanimous votes (more than 90\% on either side); cast codes are mapped to a binary scale with abstentions treated as missing.
The roll-call data serves as a binary setting with high $K$ and low $N$ that complements the \smartvote data.

\paragraph{Polis (vTaiwan).}
\polis is an online deliberation platform on which participants vote \emph{agree}, \emph{disagree}, or \emph{pass} on user-generated statements~\citep{small_polis_2021}.
We use the public vTaiwan dataset, in which 1,921 participants discussed the regulation of taxi services in Taiwan.
The original ternary codes are mapped to $\{0, 0.5, 1\}$.
The resulting response matrix is sparse by design: each participant only rates a subset of the statements, leaving 86.9\% of entries missing.
The Polis data serves as a stress test for models to fit with a large fraction of missing data.

\paragraph{European Values Study (2017).}
The European Values Study (EVS) surveys attitudes toward family, work, religion, politics, and society across European countries~\citep{evs_european_2017}.
We restrict our analysis to the numerical subset of 32 items of the EVS 2017 dataset, drawn from the Joint EVS/WVS 2017--2021 release~\citep{evs/wvs_joint_2024}.
After removing respondents with any unanswered item and restricting to countries with more than 40 respondents, we retain 35,116 respondents across 34 countries.
Each item is rescaled per question to $[0, 1]$.
This dataset tests how well the algorithms perform with large numbers of users.

\subsection{Metrics}
\label{sec:metrics}

The four-cell evaluation framework of~\citet{bachmann_fast_2024} operates on two splits of the response matrix: 
first, it splits the data into train and test users; second, it holds out a fraction of existing responses and distinguishes between reconstructing the observed and imputing the withheld answers.
Therefore, \emph{train reconstruction} evaluates how well the fitted model reconstructs the observed answers of train users.
\emph{Train imputation} evaluates how well the fitted model predicts the withheld answers of train users.
\emph{Test reconstruction} evaluates how well the fitted model generalizes to held-out users, measuring the reconstruction of their observed answers.
Finally, \emph{test imputation} evaluates how well held-out users' remaining questions can be imputed based on the embedding of their observed answers. 

\paragraph{Prediction metrics.}
Within each of the four evaluation cells, we report two metrics that emphasize complementary aspects of the prediction:
The mean absolute error (MAE) treats the prediction as a continuous agreement-probability and sums the absolute differences between predicted and true values.
The accuracy (ACC) considers a binary classification and counts a prediction as correct whenever the predicted and true responses fall on the same side of 0.5. 
While MAE rewards nuanced model predictions, ACC is more commonly used to assess model fit in ideal point estimation~\citep{bafumi_practical_2005}.
For ACC, we exclude neutral ground-truth values at exactly $y=0.5$. 

\paragraph{Latent-geometry diagnostics.}
We additionally report two diagnostics of the latent embedding.
We define the boundary fraction (BF) as the share of users within 10\% of the grid boundary; larger values indicate clusters at the boundary.
By distortion (DIS), we refer to the expected displacement of a latent point under the model parameters. 
This metric is obtained by generating uniformly distributed users at positions $x_n$ and re-embedding them based on their synthetic responses $y_k(x_n)$ using the model's item parameters.
The distortion is then computed as the Euclidean distance between the original position $x_n$ and the resulting posterior mean $\tilde{x}_n$.
To approximate the distortion, we use a $30 \times 30$ grid spanning the latent space.

\subsection{Experimental Setup}
\label{sec:experiments-design}

We run two experiments to evaluate the performance of \IXPLORE.
The first experiment compares \IXPLORE against ten state-of-the-art reference algorithms.
The second experiment analyzes \IXPLORE's configuration, varying three parameters of its optimization procedure (prior variance, number of iterations, and feature transforms).

\subsubsection{Reference Algorithms}
\label{sec:baseline-comparison}

The core question of this report is {whether \IXPLORE reconstructs and imputes responses at least as well as established latent-space models, and whether it generalizes from train users to held-out test users under increasing sparsity}.
We compare \IXPLORE against ten reference algorithms:
First, we consider PCA with linear and logistic decoding~\citep{potthoff_estimating_2018}, as well as kernel PCA paired with a per-item support vector machine~\citep{scholkopf_learning_2002}.
Second, we consider the multi-dimensional scaling variants \tSNE~\citep{vandermaaten_visualizing_2008} and \UMAP~\citep{mcinnes_umap_2018}, both with logistic decoding.
Third, we consider two VAEs~\citep{kingma_autoencoding_2014}, in a 2-layer and a logistic-decoder variant as previously used as a reference by~\citet{bachmann_fast_2024}.
Fourth, we use the IRT models \IDEAL~\citep{clinton_statistical_2004}, its expectation-maximization counterpart \emIRT~\citep{imai_fast_2016}, and \LSIRM~\citep{lee_euclidean_2025}.
We do not include \NOMINATE~\citep{poole_spatial_1985}, because its estimation does not handle response matrices with many missing values well, which rules it out for the sparsity levels and the \polis data considered here.
The exact configuration of each reference algorithm is given in \cref{app:reference-algorithms}.
\IXPLORE itself is run both with the original and binarized response matrix at $T = 10$ iterations and otherwise default parameters (see \cref{tab:hparams} in \cref{app:model-details}).

\paragraph{Predictive performance.}
Each model is fitted on the train users at each train sparsity $u \in \{0.0, 0.3, 0.6, 0.9\}$ over five random masking seeds, recording train reconstruction and imputation at every sparsity and seed.
The test users are embedded by the model fitted with unmasked train data ($u = 0$) and evaluated across the test-sparsity levels $v \in \{0.0, 0.3, 0.6, 0.9\}$, again over five seeds.
We report each metric both at every individual sparsity level and as a single aggregate over all sparsities.
While the aggregate can hide differences that appear only at specific sparsity levels, it yields one number per evaluation cell, dataset, and metric, which makes the comparison across all five datasets easier to interpret.

\paragraph{Latent dimensionality.}
All reference algorithms embed into a two-dimensional space to match \IXPLORE's 2D grid (except \emIRT, whose current implementation is constrained to one dimension).
To separate the choice of algorithm from the choice of latent dimensionality, {we further ask how many dimensions a PCA baseline needs to match \IXPLORE's predictive performance}.
To this end, we sweep PCA with linear and logistic decoding over $d \in \{1, \dots, 74\}$ dimensions, using the same sparsity levels and seeds, and read off, per evaluation cell, the smallest $d$ at which each variant reaches the reconstruction and imputation error of \IXPLORE.

\subsubsection{Configuration Analysis}
\label{sec:configuration-analysis}

The second objective of this report is to analyze how the configuration of \IXPLORE shapes its performance and latent geometry.
For these experiments, we focus on the \smartvote 2023 data.

\paragraph{Prior regularization.}
\label{sec:experiments-prior}
The prior variance sets the diagonal of the Gaussian prior on the latent space in \cref{eq:prior}, with a smaller $\tau^2$ concentrating the prior near the origin and a larger $\tau^2$ flattening it toward a uniform prior.
{We ask whether this prior variance affects the reconstruction and imputation error, as well as the latent-geometry diagnostics.}
For all values of $\tau^2 \in \{0.05,\, 0.1,\, 0.25,\, 0.5,\, 1.0,\, 10^{6}\}$, we train an \IXPLORE model for $T=20$ iterations with PCA initialization and report the prediction metrics and latent-geometry diagnostics aggregated over the sparsity levels $u \in \{0.0, 0.3, 0.6, 0.9\}$.

\paragraph{Iterative optimization.}
\label{sec:experiments-iteration}
After initialization, each iteration first refits the per-user posteriors against the current item models via \cref{eq:posterior-mean}, then refits the item models against the new posterior point estimates via \cref{eq:item-step}.
{We ask (i)~whether the iterative optimization of posteriors and item models improves the reconstruction and imputation error, and (ii)~whether PCA initialization reduces the number of iterations until convergence.}
We vary the iteration count $T \in \{0, 1, 2, 5, 10, 20\}$ and the initialization scheme (PCA vs.\ random) over the standard sparsity levels and seeds with $\tau^2 = 0.25$ fixed.
The prediction metrics and latent-geometry diagnostics are tracked as a function of $T$, and we compare how many iterations the PCA and random initializations each need to reach the same reconstruction and imputation error.

\paragraph{Feature transforms.}
\label{sec:experiments-features}
In the response function of \cref{eq:response}, the optional feature transform $\varphi$ maps each 2D latent position into a higher dimensional vector before the per-item logistic regression.
{We ask whether non-linear feature transforms improve reconstruction and imputation error while keeping decision boundaries visually interpretable.}
We compare the default identity transform with linear decision boundaries; a polynomial transform with squared coordinates, $\varphi(x) = (x_1, x_2, x_1^2, x_2^2)$; and an RFF transform that projects the position into a high-dimensional cosine basis to approximate an RBF kernel~\citep{rahimi_random_2007}.
We evaluate the three feature transforms over the standard sparsity levels and seeds at the default configuration (PCA initialization, $\tau^2 = 0.25$, $T = 10$).

\begin{figure}[ht]
    \centering
    \includegraphics{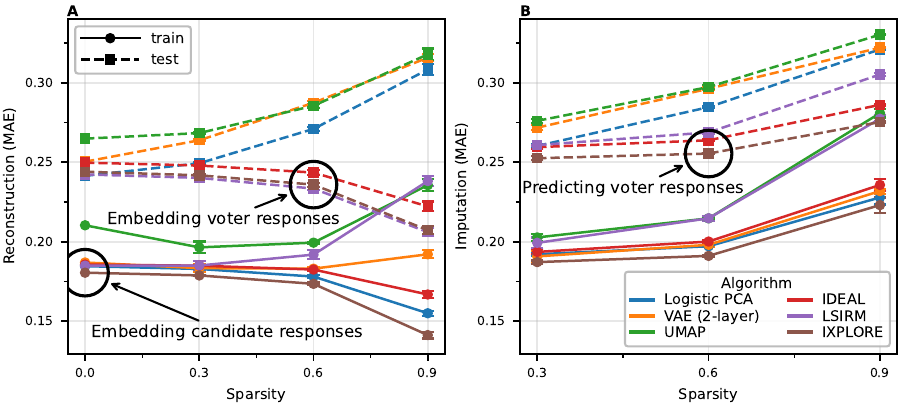}
    \caption{\IXPLORE and five reference algorithms on \smartvote 2023. Reconstruction and imputation error are reported as a function of sparsity in each of the four evaluation cells. Three points of interest are highlighted: the train reconstruction error at $u=0$ corresponds to embedding candidates with full data; the test reconstruction error (e.g., at $v=0.6$) corresponds to embedding voters in the same space; the test imputation error (e.g., at $v=0.6$) corresponds to predicting voters' remaining responses.}
    \label{fig:baseline_comparison}
\end{figure}

\section{Results}

We report our findings in two sections corresponding to the experiments described in \cref{sec:experiments-design}.
All data and code to reproduce the results are available in our repository on GitHub.\footnote{\url{https://github.com/fsvbach/ixplore-paper}}

\subsection{Reference Algorithms}
\label{sec:results-baseline}

\cref{fig:baseline_comparison} compares \IXPLORE against reference algorithms on the \smartvote 2023 data. 
As an overview, we show a selection of five reference algorithms that cover the variance across methods.
Similar figures for the other datasets are provided in \cref{app:results} (\crefrange{fig:baseline_train_test_smartvote_2019}{fig:baseline_train_test_evs}).

\paragraph{Reconstruction error.} 
Panel A shows the reconstruction error across different sparsity values.
Most algorithms reconstruct the full data at $u=0$ well, starting from an MAE between 0.181 (\IXPLORE) and 0.210 (\UMAP).
With increasing sparsity, the range of the reconstruction error grows:
\IXPLORE further reduces the MAE to 0.141, while the MAE of \LSIRM and \UMAP increases to around 0.238.
For the test data, all reconstruction errors are significantly higher due to the distribution shift from candidates (train) to voters (test).
The algorithms fall into two categories:
model-based algorithms (\IDEAL, \LSIRM, and \IXPLORE) improve performance with increasing sparsity, while the ML-based algorithms (\UMAP, VAE, and PCA) struggle to reconstruct test data at higher sparsity. 
At $v=0.9$, the difference in MAE grows to a range from 0.206 (\LSIRM) to 0.318 (\UMAP). 

\paragraph{Imputation error.} 
Panel B shows the corresponding imputation error at $v \in \{0.3, 0.6, 0.9\}$.
At low sparsity, the imputation errors are comparable across algorithms.
At $u=0.9$, the separation between algorithms becomes more visible:
\LSIRM and \UMAP increase the imputation error up to 0.281, while \IXPLORE's error remains around 0.223.
For the test data, the distinction between model-based and ML-based algorithms is less visible than before.
Notably, for both training and testing, \IXPLORE reaches the lowest imputation error across all sparsities.

\begin{table}[ht]
\centering
\caption{Performance comparison on Smartvote (2023). Each cell is the mean over all sparsity levels. Reference algorithms are listed in the top block; \IXPLORE is run both with the continuous input and the binarized input$^*$. The best-performing algorithm is shown in \textbf{bold}, the runner-up in \textit{italics}.}
\label{tab:baseline_smartvote_2023}
\small
\begin{tabular}{lcccccccc}
\toprule
 & \multicolumn{2}{c}{Train Rec.} & \multicolumn{2}{c}{Train Imp.} & \multicolumn{2}{c}{Test Rec.} & \multicolumn{2}{c}{Test Imp.} \\
\cmidrule(lr){2-3}\cmidrule(lr){4-5}\cmidrule(lr){6-7}\cmidrule(lr){8-9}
Algorithm & MAE & ACC & MAE & ACC & MAE & ACC & MAE & ACC \\
\midrule
Linear PCA & 0.245 & 80.2\% & 0.274 & 76.4\% & 0.282 & 72.4\% & 0.301 & 69.1\% \\
Logistic PCA & 0.175 & 85.1\% & \textit{0.206} & \textit{80.9\%} & 0.267 & 72.7\% & 0.289 & 69.5\% \\
Kernel PCA & 0.195 & 84.7\% & 0.223 & 80.4\% & 0.283 & 70.6\% & 0.304 & 67.4\% \\
t-SNE & 0.212 & 80.9\% & 0.239 & 76.7\% & 0.280 & 70.0\% & 0.300 & 67.0\% \\
UMAP & 0.210 & 81.1\% & 0.233 & 77.4\% & 0.284 & 69.6\% & 0.301 & 67.0\% \\
VAE (2-layer) & 0.186 & 84.5\% & 0.207 & \textbf{81.4\%} & 0.279 & 72.1\% & 0.297 & 69.0\% \\
VAE (logistic) & \textit{0.175} & 85.0\% & 0.208 & 80.2\% & 0.273 & 71.9\% & 0.293 & 68.9\% \\
IDEAL & 0.180 & 84.6\% & 0.210 & 80.1\% & 0.241 & 76.0\% & \textit{0.270} & 71.3\% \\
emIRT & 0.205 & 81.3\% & 0.221 & 79.0\% & 0.259 & 74.4\% & 0.287 & 70.8\% \\
LSIRM & 0.200 & 84.0\% & 0.230 & 78.8\% & \textbf{0.230} & \textbf{79.2\%} & 0.278 & 71.3\% \\
\midrule
IXPLORE & \textbf{0.168} & \textit{85.2\%} & \textbf{0.200} & 80.6\% & \textit{0.232} & 77.1\% & \textbf{0.261} & \textbf{72.3\%} \\
IXPLORE$^*$ & 0.177 & \textbf{86.2\%} & 0.219 & 80.0\% & 0.235 & \textit{77.9\%} & 0.270 & \textit{72.3\%} \\
\bottomrule
\end{tabular}
\end{table}

\subsubsection{Predictive Performance}

\cref{tab:baseline_smartvote_2023} aggregates MAE and ACC per evaluation cell across all sparsity levels and random seeds for the \smartvote 2023 data. 
For the other datasets, the corresponding tables are provided in \cref{app:results} (\cref{tab:baseline_smartvote_2019}--\cref{tab:baseline_evs}).

\paragraph{Comparison across metrics.}
On \smartvote 2023, \IXPLORE attains the lowest aggregated MAE in three of the four evaluation cells: train reconstruction (0.168), train imputation (0.200), and test imputation (0.261).
On test reconstruction it is the runner-up at 0.232 behind \LSIRM (0.230).
On ACC, the \IXPLORE variant trained on binarized data classifies the train data 1 percentage point better than standard \IXPLORE.
\LSIRM generalizes best to test data, leading \IXPLORE by 1.3 percentage points on test reconstruction.
For test imputation, both \IXPLORE variants classify the held-out responses equally well, ahead of the runner-up \IDEAL by 1 percentage point.

\paragraph{Comparison across datasets.}
The observed ranking is stable across all five datasets: 
In every evaluation cell, \IXPLORE achieves either the lowest or second-lowest MAE.
In particular, \IXPLORE reaches the lowest train reconstruction MAE everywhere (0.032 on \voteview up to 0.173 on EVS) and the lowest test imputation MAE on every dataset except \voteview, where it ties \IDEAL (0.083).
The margin to the runner-up is widest on \smartvote 2023 and \polis and narrowest on \voteview, where the IRT models \IDEAL and \emIRT remain competitive.
On EVS, \IXPLORE generalizes best to test data, with a train reconstruction MAE of 0.173 and a test reconstruction MAE of 0.180.
However, on this larger dataset, \IDEAL and \LSIRM could not be run due to computational constraints.

\subsubsection{Latent Dimensionality}
\label{sec:results-dimensions}

\begin{figure}[t]
    \centering
    \includegraphics{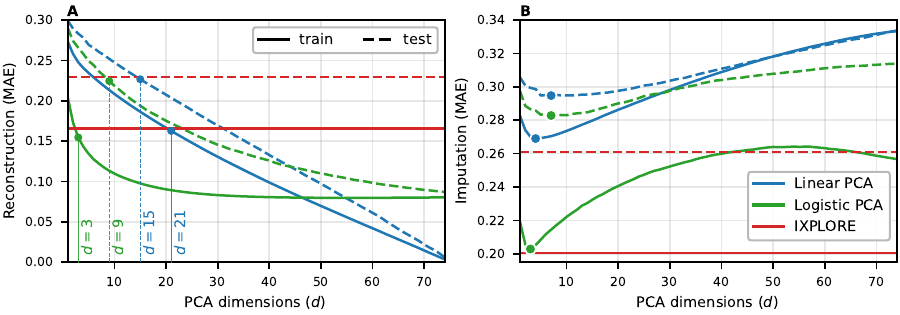}
    \caption{Reconstruction and imputation error of PCA on \smartvote 2023 as a function of the latent dimensionality $d$, averaged over all sparsity levels. The circles mark each curve's minimum. The horizontal red lines indicate the \IXPLORE performance with $\tau^2 = 0.25$, PCA initialization, and $T = 10$ iterations. {Logistic PCA} exceeds \IXPLORE's training reconstruction error for $d \geq 3$ (Linear PCA for $d \geq 21$); neither variant reaches \IXPLORE's training or test imputation error at any $d \leq 74$.}
    \label{fig:pca_dimensionality}
\end{figure}

\cref{fig:pca_dimensionality} shows the reconstruction and imputation error of the two PCA variants as a function of the latent dimensionality $d$.
The training and test reconstruction error decreases monotonically with $d$ for both variants.
The logistic variant reaches the same train reconstruction error as \IXPLORE at $d = 3$ ($d=9$ for test) and the linear variant at $d = 21$ ($d=15$ for test).
PCA's train and test imputation errors, in contrast, attain minima at small dimensionalities: 
the logistic variant decreases until $d = 3$ with a train MAE of 0.203; the linear variant decreases until $d = 4$ with a train MAE of 0.269.
For test imputation error, both minima are reached slightly later at $d = 5$ (logistic) and $d = 7$ (linear).
At none of these minima does either variant reach \IXPLORE's training (0.200) or test (0.261) imputation error.

\subsection{Configuration Analysis}
\label{sec:results-configuration}

\begin{figure}[t]
    \centering
    \includegraphics{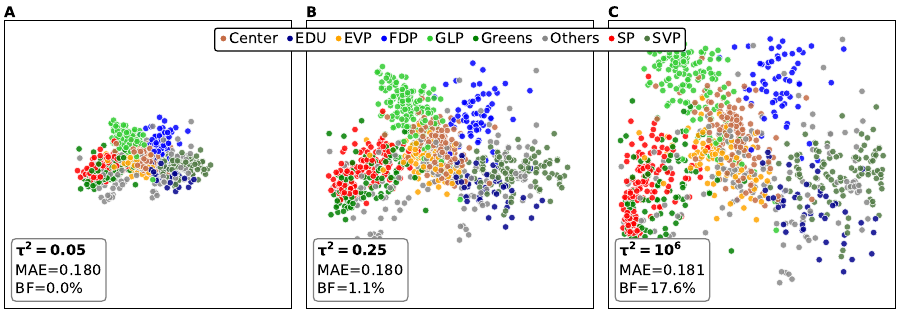}
    \caption{\IXPLORE embeddings of the full \smartvote 2023 candidate data at $T = 20$ for three different priors. The reconstruction error stays stable, while the boundary fraction (share of points within 10\% of the grid boundary) increases with $\tau^2$.}
    \label{fig:prior_embeddings}
\end{figure}

\cref{fig:prior_embeddings} shows the \IXPLORE embeddings of the \smartvote 2023 data for three different prior variances.
The tightest prior with $\tau^2=0.05$ collapses the candidates into a tight cluster around the origin (Panel A), while the flat prior with $\tau^2=10^6$ leaves 17.6\% of candidates within 10\% of the grid boundary (Panel C).
The prior at $\tau^2 = 0.25$ produces a visible party structure without large boundary clusters (Panel B). 
Notably, there is no difference in predictive performance across priors, as all embeddings reach a reconstruction MAE of around 0.18.

\begin{figure}[t]
    \centering
    \includegraphics{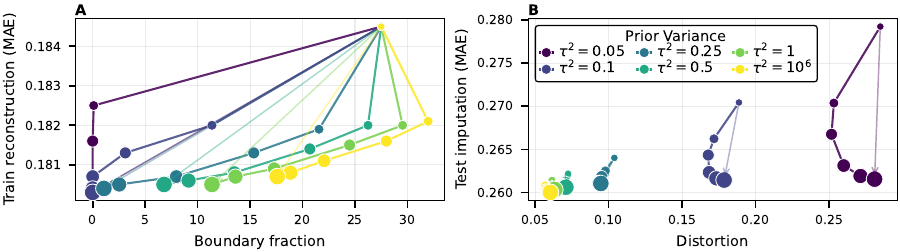}
    \caption{Iteration trajectories $T \in \{0,1,2,5,10,20\}$ for each prior $\tau^2$ with PCA initialization. The marker size grows with $T$ while the first and last iteration are connected with an arrow pointing from $T = 0$ to $T = 20$. \abc{A} The boundary fraction and train reconstruction error at $u = 0$ decrease over the course of iterations. \abc{B} The distortion of the latent space remains stable over the course of iterations, while the test imputation error decreases (averaged over $v \in \{0.3, 0.6, 0.9\}$).}
    \label{fig:prior_phase}
\end{figure}

\subsubsection{Prior Regularization}
\label{sec:results-prior}

\cref{fig:prior_phase} traces each prior's iteration trajectory from $T = 0$ to $T=20$ in a more detailed analysis.
Panel A shows the boundary fraction and train reconstruction error at $u = 0$.
At $T = 0$ the PCA initialization places every prior at the same point. 
As $T$ increases, each prior moves toward lower reconstruction error and boundary fraction at the bottom left.
Smaller priors reduce the boundary fraction much faster than larger priors, while all priors converge to a similar reconstruction error.
Panel B shows the distortion and test imputation error aggregates across sparsity levels $v\in\{0.3,0.6,0.9\}$.
While large priors already achieve minimal imputation error after initialization, smaller priors reach a similar performance only after several iterations.
The distortion of the embedding, however, is stable across iterations.
Larger priors have vanishing distortion independent of the iterations, while smaller priors first slightly decrease and then increase distortion with $T$.

\begin{figure}[ht]
    \centering
    \includegraphics{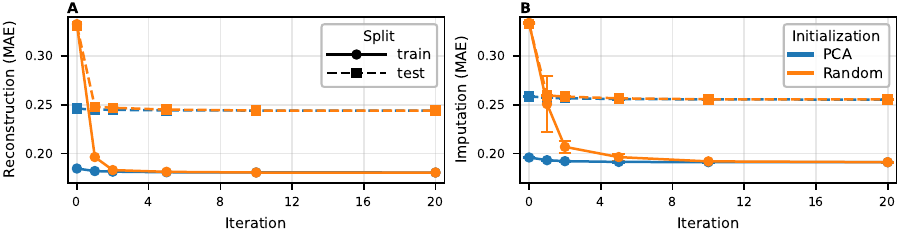}
    \caption{Effect of initialization across iterations on \smartvote 2023 at $\tau^2 = 0.25$. \abc{A} The reconstruction MAE is reported at $u = 0$. \abc{B} The imputation MAE is averaged over $v \in \{0.3, 0.6, 0.9\}$.}
    \label{fig:iteration_effect}
\end{figure}

\subsubsection{Iterative Optimization}
\label{sec:results-iteration}

\cref{fig:iteration_effect} shows the convergence of the MAE with $\tau^2=0.25$ for random and PCA initialization.
Random initialization starts with a reconstruction MAE above 0.330 and then quickly decreases to 0.183 after two iterations.
For the imputation MAE, convergence takes roughly ten iterations.
Initializing \IXPLORE with a PCA embedding significantly reduces the number of iterations needed for convergence.
With PCA initialization, \IXPLORE converges in less than two iterations, with the test-data MAE improving only minimally thereafter.
Both random and PCA initialization, however, converge to the same reconstruction and imputation errors.

\subsubsection{Feature Transforms}
\label{sec:results-features}

\begin{figure}[t]
    \centering
    \includegraphics{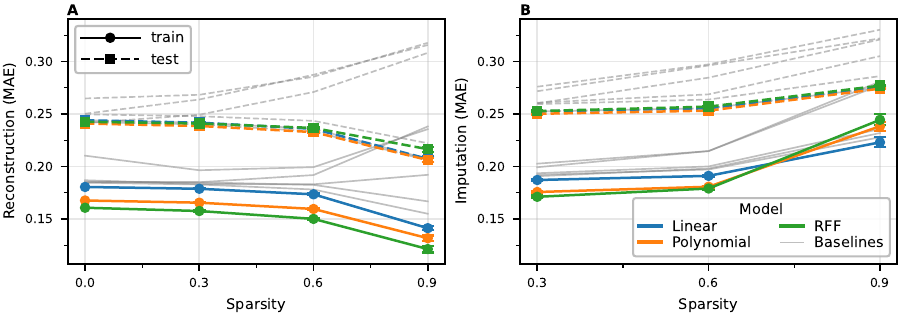}
    \caption{Reconstruction and imputation error on \smartvote 2023 data for the three \IXPLORE feature transforms (colored) against the reference algorithms (gray).}
    \label{fig:kernel_comparison}
\end{figure}

\cref{fig:kernel_comparison} reports reconstruction and imputation error for the three feature transforms compared to the reference algorithms.
Outperforming the results of the linear baseline, both the polynomial and RFF transforms can significantly improve the train reconstruction error across all sparsities.
The RFF transform, in particular, reduces the MAE from 0.166 to 0.144 on average.
The train imputation performance at high sparsities, however, is best for the linear model.
At $u=0.9$, both polynomial and RFF transforms exceed the MAE of the linear model by 0.015 and 0.021.
For test data, there is no visible difference across feature transforms.

\begin{figure}[ht]
    \centering
    \includegraphics{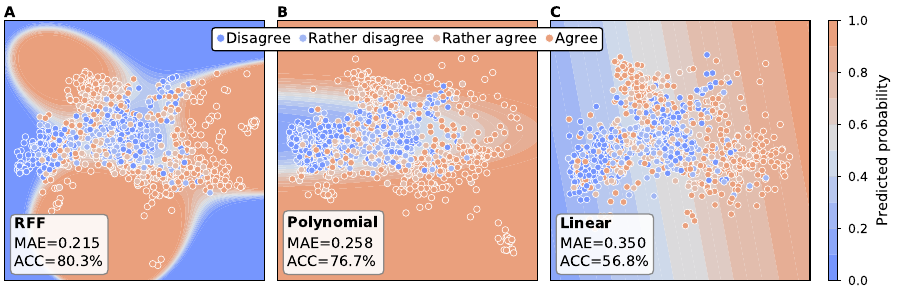}
    \caption{Predicted probability maps for three different feature transformations of the \IXPLORE model initialized with $\tau^2=0.25$ after $T=10$ iterations on the \smartvote 2023 data. In each panel, the item with the largest accuracy spread across feature transforms is chosen (item 32263). The dots correspond to the 1,029 candidates colored by their response to this item (orange/blue $=$ no/yes).}
    \label{fig:kernels_visual}
\end{figure}

In \cref{fig:kernels_visual}, we further inspect the \emph{visual interpretability} of the three feature transforms by plotting the predicted probability contours for item 32263\footnote{Question: \enquote{Should it be possible to hold a referendum on federal spending above a certain amount?}}.
This item was chosen as an example as it maximizes the range of predictive accuracy across feature transforms:
the RFF transform reaches the highest accuracy with 80.3\%, while the polynomial transform reaches 67\% and the linear baseline stays at 56\%.
The RFF surface in Panel A, however, splits the latent space into three distinct areas of agreement, which might make it harder to interpret.
The polynomial transform in Panel B, in contrast, follows the candidate clusters with a single, smooth decision boundary on the 2D latent space, which might be more desirable for visual interpretability.
The linear transform in Panel C fails to capture the non-linear structure of the data.
\section{Discussion}
\label{sec:discussion}

Our experimental results show that \IXPLORE matches or exceeds state-of-the-art spatial models in terms of reconstruction and imputation quality.
Aggregated across sparsities, \emph{\IXPLORE attains the lowest or second-lowest MAE in every evaluation cell across the five datasets}.
We discuss what drives this performance, how \IXPLORE's configuration shapes the embedding geometry, and what these findings imply for spatial modeling of political data.

\subsection{Predictive Performance}

Regardless of the number of latent dimensions, PCA cannot match \IXPLORE's imputation error on sparse users in the \smartvote 2023 data (see~\cref{fig:pca_dimensionality}).
This performance gap cannot be explained by the iterative optimization alone, which does \emph{not} dramatically increase \IXPLORE's predictive performance over the PCA initialization.
Instead, we attribute the improved performance to \IXPLORE's inference mechanism:
while ML-based models---such as PCA---minimize the reconstruction error, they must impute missing responses \emph{before} embedding a new user.
\IXPLORE avoids this step through a sparsity-aware likelihood function inspired by IRT models.
For sparse test users in particular, this framework generalizes well and outperforms even \IDEAL, which was previously found to excel at modeling the \smartvote data~\citep{bachmann_fast_2024}.
This performance gain may stem from the fact that IRT-based models---such as \IDEAL---are derived from a spatial-utility function rather than optimized to predict held-out responses~\citep{clinton_simulate_2009}.

However, the results also showed some indications of overfitting.
While the generalization gap between train and test reconstruction is small for \IXPLORE (comparable to IRT baselines and much smaller than for ML baselines), the gap between train reconstruction and train imputation is larger for \IXPLORE on binary data.
On both \voteview and \polis, this reconstruction-imputation gap exceeds the corresponding gap of \IDEAL, especially at high sparsity.
This is likely driven by the weak L2 regularization used in our experiments for \IXPLORE, which lets the per-item logistic regression fit the few observed binary responses too tightly.
At the same time, this regularization choice also contributes to \IXPLORE's lowest train reconstruction on ordinal data, where the continuous response values bound the logistic gradient and provide implicit regularization.

\subsection{Latent Geometry}

The choice of the prior variance does not affect the predictive quality of \IXPLORE.
This indicates that the prior acts purely as a geometric regularizer, controlling the trade-off between boundary fraction and distortion of the embedding.
This regularization is particularly important for \IXPLORE's bounded latent space, as the Bernoulli likelihood pushes the posterior mass outwards for users with consistent response profiles.
A tight prior pulls the embedding back toward the origin, however, at the cost of a higher distortion.
Whether or not this distortion is problematic depends on the application:
the higher the distortion, the more difficult it becomes for users to reach the extremes of the latent space, possibly undermining the credibility of the latent geometry.
An intermediate prior variance balances these forces, letting the embedding contract while keeping a low boundary fraction.
The optimal value for $\tau^2$, however, must be identified by visual inspection, which makes prior selection a target for automation in future work.

\subsection{Non-Linear Feature Transforms}

Non-linear feature transforms can significantly improve the predictive performance of \IXPLORE.
The RFF transform, inspired by the RBF kernel~\citep{scholkopf_learning_2002,rahimi_random_2007}, outperforms both the linear and polynomial transforms in terms of reconstruction and imputation error.
Choosing such a non-linear transform, however, departs from the tradition of utility functions in ideal point estimation~\citep{eguia_challenges_2013}.
While the logistic regression of the linear \IXPLORE variant resembles the quadratic utility function of IRT models, the non-linear transforms do not have a straightforward interpretation in terms of spatial utility.
In this trade-off between accuracy and interpretability, the polynomial transform (\cref{fig:kernels_visual}B) may be a reasonable compromise, as its contour follows the candidate clusters with a single smooth decision boundary.

\subsection{Implications}

\IXPLORE is well suited for applications in which visualization is at the core of the analysis, since its bounded latent space keeps all users within a fixed frame.
It is particularly useful when predictive performance should be combined with visual interpretability, either through the per-item logistic regression decision boundaries, or through the grid-based uncertainty quantification\footnote{On \smartvote 2023, \IXPLORE locates half of the held-out users within their final $1\sigma$ posterior ellipse after roughly a third of the items (see \cref{app:posterior-convergence}).}.
\IXPLORE is also suited for online applications where inference must be fast: held-out users are embedded through a single grid posterior evaluation against fixed item parameters.
Beyond these direct use cases, \IXPLORE fills a methodological niche as one of few Python-based IRT implementations, offering an alternative to the R-based tooling that has historically dominated the field.
For developers of opinion landscapes and analysts of political response data, \IXPLORE therefore offers a new option in the algorithmic toolbox.

\section{Limitations}
\label{sec:limitations}

There are several limitations to the current implementation of \IXPLORE, as well as to the experimental setup used to evaluate its performance.

\paragraph{Implementation.}
With $R^d$ grid points for a $d$-dimensional latent space, \IXPLORE's grid-based posterior inference does not scale beyond two dimensions.
Moreover, the grid-based inference requires matrix multiplications over $R^2 \times N$ entries per iteration, which becomes prohibitive for very large user populations.
Extending \IXPLORE to such regimes---through sparse or hierarchical grid approximations---is a natural direction for future work.
Furthermore, as with other dimensionality-reduction algorithms, the axes of the \IXPLORE embedding do not carry an inherent meaning.
When visualizing the embedding, this requires manual labeling of the axes.
In contrast to other algorithms, \IXPLORE embeddings are not rotation invariant: the square grid breaks the rotational symmetry of the likelihood, so two runs with random initialization cannot be aligned by a post-hoc rotation as in models such as \IDEAL.
Future work could explore the use of language models to automatically label the axes based on the most influential items along each.

\paragraph{Evaluation.}
The evaluation criteria chosen in this report almost exclusively reward predictive performance.
Yet for a political map intended for interpretation and visualization, reconstruction and imputation accuracy may not be the only objective.
Other criteria commonly used to evaluate ideal point estimation algorithms---such as the unidimensionality, reliability, and data quality of the derived scales~\citep{bruinsma_comparison_2020}---are not considered in this report.
Furthermore, while the chosen datasets for \IXPLORE's evaluation cover a wide range of applications and datatypes, some of the reference algorithms only take binary values (e.g., \IDEAL).
This was shown to be an advantage when optimizing for accuracy but a disadvantage for the reconstruction error, as it prevents them from fitting the ordinal responses.
Finally, the sparsity in the data is induced synthetically: apart from \polis, all datasets are complete, and we simulate sparse responders by masking answers uniformly at random, which does not reproduce the selective non-response of real users in real-world applications.
Within the scope of our evaluation, however, \IXPLORE proves to be a robust and effective approach to latent space modeling.

\section{Conclusion}
\label{sec:conclusion}

We presented \IXPLORE, an ideal point estimation algorithm that places users on a bounded 2D latent space through grid-based posterior inference.
Existing methods force a choice between model-based algorithms, which are not optimized to predict held-out responses, and ML-based algorithms, which generalize poorly to sparse users.
\IXPLORE combines the strengths of both families with a sparsity-aware likelihood function and a predictive fit objective.
Our experiments across five datasets showed that \emph{this approach matches or surpasses both model-based and ML-based models on reconstruction and imputation error}.
This is particularly relevant for the sparse responses of online deliberation settings and the dense ordinal data of VAAs.

Beyond predictive performance, we studied how \IXPLORE's configuration shapes the geometry of the latent space.
We found that the \emph{prior variance trades boundary fraction against distortion without changing the reconstruction error}, while iterations further refine the embedding with slight improvement in imputation error.
When using \IXPLORE, we recommend PCA initialization for reproducibility and faster convergence.
As an extension to standard ideal point estimation, non-linear feature transforms increase the flexibility of \IXPLORE: the RFF transform significantly lowered the train reconstruction error, while the polynomial transform kept the decision boundaries visually interpretable.

Based on these findings, we conclude that \IXPLORE offers a novel framework for constructing predictive political maps.
Several of its components---the bounded latent space, the grid-based posterior, and the per-item response surfaces---are \emph{directly tailored to visualization settings} in which new users arrive with sparse answers.
In interactive applications such as adaptive questionnaires, VAAs, and deliberation platforms, where users must be embedded online and their remaining responses predicted, \IXPLORE provides a competitive alternative to both classical ideal point models and generic ML embeddings.
Thereby, our study contributes to visualizing political maps as instruments for democratic education.

\section*{Acknowledgements}

I acknowledge the Swiss National Science Foundation (SNSF), which funds the research with grant ID CRSII5-205975. 
Furthermore, I am very grateful to the team from \textit{Politools} for providing the \smartvote data. 
Lastly, I thank my supervisors, collaborators, and students for their insights and support throughout the project.

\bibliography{AdjustedLibrary}

@article{bachmann_adaptive_2025,
  title = {Adaptive political surveys and {{GPT-4}}: {{Tackling}} the cold start problem with simulated user interactions},
  shorttitle = {Adaptive political surveys and {{GPT-4}}},
  author = {Bachmann, Fynn and {van der Weijden}, Daan and Heitz, Lucien and Sarasua, Cristina and Bernstein, Abraham},
  year = 2025,
  month = may,
  journal = {PLOS One},
  volume = {20},
  number = {5},
  pages = {e0322690},
  urldate = {2025-06-06},
  copyright = {All rights reserved},
  langid = {english}
}

@article{bachmann_estimating_2026,
  title = {Estimating the {{Recommendation Certainty}} in {{Candidate}}-{{Based Voting Advice Applications}}},
  author = {Bachmann, Fynn and {van der Weijden}, Daan and Sarasua, Cristina and Bernstein, Abraham},
  year = 2026,
  month = jan,
  journal = {Politics and Governance},
  volume = {14},
  number = {0},
  pages = {23},
  urldate = {2026-01-21},
  copyright = {All rights reserved},
  langid = {english}
}

@incollection{bachmann_fast_2024,
  title = {Fast and {{Adaptive Questionnaires}} for {{Voting Advice Applications}}},
  booktitle = {Machine {{Learning}} and {{Knowledge Discovery}} in {{Databases}}. {{Applied Data Science Track}}},
  author = {Bachmann, Fynn and Sarasua, Cristina and Bernstein, Abraham},
  year = 2024,
  month = aug,
  volume = {14950},
  pages = {365--380},
  publisher = {Springer Nature Switzerland},
  address = {Cham},
  urldate = {2024-09-03},
  copyright = {All rights reserved},
  langid = {english}
}

@incollection{bachmann_wasserstein_2023,
  title = {Wasserstein t-{{SNE}}},
  booktitle = {Machine {{Learning}} and {{Knowledge Discovery}} in {{Databases}}},
  author = {Bachmann, Fynn and Hennig, Philipp and Kobak, Dmitry},
  year = 2023,
  month = mar,
  volume = {13713},
  pages = {104--120},
  publisher = {Springer International Publishing},
  address = {Cham},
  urldate = {2024-02-23},
  copyright = {All rights reserved},
  langid = {english}
}

@article{bafumi_leapfrog_2010,
  title = {Leapfrog {{Representation}} and {{Extremism}}: {{A Study}} of {{American Voters}} and {{Their Members}} in {{Congress}}},
  shorttitle = {Leapfrog {{Representation}} and {{Extremism}}},
  author = {Bafumi, Joseph and Herron, Michael C.},
  year = 2010,
  month = aug,
  journal = {American Political Science Review},
  volume = {104},
  number = {3},
  pages = {519--542},
  urldate = {2023-01-17},
  langid = {english}
}

@article{bafumi_practical_2005,
  title = {Practical {{Issues}} in {{Implementing}} and {{Understanding Bayesian Ideal Point Estimation}}},
  author = {Bafumi, Joseph and Gelman, Andrew and Park, David K. and Kaplan, Noah},
  year = 2005,
  month = mar,
  journal = {Political Analysis},
  volume = {13},
  number = {2},
  pages = {171--187},
  urldate = {2026-02-01},
  copyright = {https://www.cambridge.org/core/terms},
  langid = {english}
}

@article{barbera_birds_2015,
  title = {Birds of the {{Same Feather Tweet Together}}: {{Bayesian Ideal Point Estimation Using Twitter Data}}},
  shorttitle = {Birds of the {{Same Feather Tweet Together}}},
  author = {Barber{\'a}, Pablo},
  year = 2015,
  month = jan,
  journal = {Political Analysis},
  volume = {23},
  number = {1},
  pages = {76--91},
  urldate = {2025-08-26},
  copyright = {https://www.cambridge.org/core/terms},
  langid = {english}
}

@article{boche_new_2018,
  title = {The new {{Voteview}}.com: preserving and continuing {{Keith Poole}}'s infrastructure for scholars, students and observers of {{Congress}}},
  shorttitle = {The new {{Voteview}}.com},
  author = {Boche, Adam and Lewis, Jeffrey B. and Rudkin, Aaron and Sonnet, Luke},
  year = 2018,
  month = jul,
  journal = {Public Choice},
  volume = {176},
  number = {1-2},
  pages = {17--32},
  urldate = {2023-09-25},
  langid = {english}
}

@article{bruinsma_comparison_2020,
  title = {A comparison of measures to validate scales in voting advice applications},
  author = {Bruinsma, Bastiaan},
  year = 2020,
  month = aug,
  journal = {Quality \& Quantity},
  volume = {54},
  number = {4},
  pages = {1299--1316},
  urldate = {2025-06-17},
  langid = {english}
}

@article{carroll_comparing_2009,
  title = {Comparing {{NOMINATE}} and {{IDEAL}}: {{Points}} of {{Difference}} and {{Monte Carlo Tests}}},
  shorttitle = {Comparing {{NOMINATE}} and {{IDEAL}}},
  author = {Carroll, Royce and Lewis, Jeffrey B. and Lo, James and Poole, Keith T. and Rosenthal, Howard},
  year = 2009,
  month = nov,
  journal = {Legislative Studies Quarterly},
  volume = {34},
  number = {4},
  pages = {555--591},
  urldate = {2023-01-16},
  langid = {english}
}

@article{clinton_simulate_2009,
  title = {To {{Simulate}} or {{NOMINATE}}?},
  author = {Clinton, Joshua D. and Jackman, Simon},
  year = 2009,
  month = nov,
  journal = {Legislative Studies Quarterly},
  volume = {34},
  number = {4},
  pages = {593--621},
  urldate = {2025-09-04},
  langid = {english}
}

@article{clinton_statistical_2004,
  title = {The {{Statistical Analysis}} of {{Roll Call Data}}},
  author = {Clinton, Joshua D. and Jackman, Simon and Rivers, Douglas},
  year = 2004,
  month = may,
  journal = {American Political Science Review},
  volume = {98},
  number = {2},
  pages = {355--370},
  urldate = {2023-01-17},
  langid = {english}
}

@incollection{costalobo_mapping_2010,
  title = {Mapping the {{Political Landscape}}: {{A Vote Advice Application}} in {{Portugal}}},
  shorttitle = {Mapping the {{Political Landscape}}},
  booktitle = {Voting advice applications in {{Europe}}: the state of the art},
  author = {Costa Lobo, Marina and Vink, Maarten P. and Lisi, Marco},
  year = 2010,
  month = oct,
  pages = {143--185},
  publisher = {Social Science Research Network},
  urldate = {2025-09-08},
  langid = {english}
}

@incollection{eguia_challenges_2013,
  title = {Challenges to the {{Standard Euclidean Spatial Model}}},
  booktitle = {Advances in {{Political Economy}}},
  author = {Eguia, Jon X.},
  year = 2013,
  pages = {169--180},
  publisher = {Springer Berlin Heidelberg},
  address = {Berlin, Heidelberg},
  urldate = {2026-05-29},
  langid = {english}
}

@misc{evs_european_2017,
  title = {European {{Values Study}} 2017: {{Integrated Dataset}} ({{ZA7500}}, {{Version}} 4.0.0)},
  author = {{EVS}},
  year = 2017,
  publisher = {GESIS Data Archive, Cologne. ZA7500 Data file},
}

@misc{evs/wvs_joint_2024,
  title = {{Joint EVS/WVS 2017-2022 Dataset (Joint EVS/WVS)}},
  author = {{EVS/WVS}},
  year = 2024,
  publisher = {GESIS Data Archive},
  urldate = {2026-05-17},
  collaborator = {Gedeshi, Ilir and Pachulia, Merab and Rotman, David and Kritzinger, Sylvia and Poghosyan, Gevorg and Fotev, Georgy and {Kolenovi{\'c}-{\DJ}apo}, Jadranka and Baloban, Stjepan and Baloban, Josip and Rabu{\v s}ic, Ladislav and Frederiksen, Morten and Saar, Erki and Ketola, Kimmo and Br{\'e}chon, Pierre and Wolf, Christof and Rosta, Gergely and Voas, David and Rovati, Giancarlo and J{\'o}nsd{\'o}ttir, Gu{\dh}bj{\"o}rg A. and Ziliukaite, Ruta and Petkovska, Antoanela and Reeskens, Tim and Komar, Olivera and Jenssen, Anders T. and Voicu, Bogdan and Soboleva, Natalia and Marody, Miros{\l}awa and Be{\v s}i{\'c}, Milo{\v s} and Strapcov{\'a}, Katarina and Uhan, Samo and Silvestre Cabrera, Mar{\'i}a and {Wallman-Lund{\aa}sen}, Susanne and Ernst St{\"a}hli, Mich{\`e}le and Ramos, Alice and Mic{\'o} Ib{\'a}{\~n}ez, Joan and Carballo, Marita and McAllister, Ian and Bangladesh), Roberto Stefan (PI, Foa and Moreno Morales, Daniel E. and De Oliveira De Castro, Henrique Carlos and Lagos, Marta and Zhong, Yang and Colombia), Andres (PI, Casas and Cyprus), Birol (PI, Yesilada and Paez, Cristina and Abdel Latif, Abdel Hamid and Ethiopia), Will (PI, Jennings and Welzel, Christian and D{\'i}az Argueta, Julio C{\'e}sar and Cheng, Edmund and Indonesia), Timothy (PI, Gravelle and Stoker, Gerry and Dagher, Munqith and Yamazaki, Seiko and Braizat, Fares and Rakisheva, Botagoz and Bakaloff, Yuri and Lebanon), Christian (PI, Haerpfer and {Wing-Yat Yu}, Eilo and Lee, Grace and Moreno, Alejandro and Souvanlasy, Chansada and Perry, Paul and Nicaragua), Carlos (PI, Denton and Nigeria), Bi (PI, Puranen and Gilani, Bilal and Romero, Catalina and Guerrero, Linda and Hern{\'a}ndez Acosta, Javier J. and Zavadskaya, Margarita and Veskovic, Nino and Auh, Soo Young and Tsai, Ming-Chang and Olimov, Muzaffar and Bureekul, Thawilwadee and Ben Hafaiedh, Abdelwahab and Esmer, Yilmaz and Inglehart, Ronald and Depouilly, Xavier and Zimbabwe), Pippa (PI, Norris and Balakireva, Olga and {Koniordos. Sokratis} and {EVS 2017:Center For Economic And Social Studies (CESS), Tirana, Albania} and {InterRating CoLtd, Yerevan, Armenia} and {Institut F\"ur Empirische Sozialforschung (IFES) GmbH, Vienna, Austria} and {Sorgu, Baku, Azerbaijan} and {Centre For Sociological And Political Research, Belarusian State University, Minsk, Belarus} and {Custom Concept D.O.O., Sarajevo, Bosnia And Herzegovina} and {Alpha Research LTD, Sofia, Bulgaria} and {Catholic University Of Croatia, Zagreb, And GfK Research Agency, Zagreb, Croatia} and {STEM/MARK, A.S., Praha, Czech Republic} and {Statistics Denmark-Survey, Copenhagen, Denmark} and {AS Emor,Tallinn, Estonia} and {Taloustutkimus Oy, Lemuntie 9, 00910 Helsinki, Finland} and {KANTAR PUBLIC-TAYLOR NELSON SOFRES, Paris, France} and {GORBI (Georgian Opinion Research Business International), Tbilisi, Georgia} and {Kantar Deutschland GmbH, Kantar Public, M\"unchen, Germany} and {NatCen Social Research, London, Great Britain} and {Forsense, Budapest, Hungary} and {Social Science Research Institute, SSRI, University Of Iceland, Reykjavik, Iceland} and {Doxa Spa, Milano, Italy} and {Baltic Surveys, Vilnius, Lithuania} and {DeFacto Consultancy, Podgorica, Montenegro} and {I\&O Research B.V., Enschede, Netherlands AndCentERdata, Tilburg, Netherlands} and {Faculty Of Philosophy, Skopje, North Macedonia} and {Statistics Norway, Oslo, Norway} and {Centrum Badania Opinii Spo\l ecznej (Public Opinion Research Centre), Warszawa, Poland} and {IRES: Institutul Roman Pentru Evaluare Si Strategie, Romania} and {CESSI (Institute For Comparative Social Research), Moscow, Russia} and {Nina Media, Novi Sad, Serbia} and {Kantar TNS, Bratislava, Slovakia} and {University Of Ljubljana, Faculty Of Social Science, Ljubljana, Slovenia} and {MyWord Research SL, Madrid, Spain} and {IPSOS Observer Sweden AB, H\"arn\"osand, Sweden} and {M.I.S Trend S.A} and {Lausanne, Switzerland (Face-To-Face) AndSwiss Centre For Expertise In The Social Sciences FORS C/O University Of Lausanne, Lausanne, Switzerland (Web-Mail)} and {GfK-Metris, Lisbon, Portugal.WVS Wave 7Institut D'Estudis Andorrans, Centre De Recerca Sociol\`ogica (CRES), Andorra} and {Voices Research And Consultancy S.A., Argentina} and {Centre For Social Research And Methods, Australian National University} and {SRG Bangladesh Limited (SRGB), Bangladesh} and {CIUDADANIA, Comunidad De Estudios Sociales Y Acci\'on P\'ublica, Bolivia} and {Federal University Of Rio Grande Do Sul, Brazil} and {Market Opinion Research International, Chile} and {Public Opinion Research Center Of School Of International And Public Affairs At Shanghai Jiao Tong University, China} and {Invamer, Colombia} and {Cymar Research Company (Survey In Cyprus South)} and {Prologue Consulting Ltd. (Survey In Cyprus North)} and {IPSOS Ecuador} and {Egyptian Research And Training Center, Egypt} and {WAAS International/ TNS RMS Nigeria Limited / Kantar (Survey In Ethiopia)} and {National Centre Of Social Research (EKKE) \& DIANEOSIS \& Metron Analysis, Greece} and {Innovation, Development \& Research, S.A. \& Social Work School Of The University Of San Carlos Of Guatemala} and {Centre For Communication And Public Opinion Survey (CCPOS) Of The Chinese University Of Hong Kong For FTF} and {Survey Sampling International (SSI) For CAWI, Hong Kong SAR PRC} and {Survey Meter, Indonesia} and {R-Research Limited, UK (Survey In Iran)} and {International Institute For Administration And Social Survey (IIACSS), Jordan (Survey In Iraq)} and {Nippon Research Center, Ltd., Japan} and {NAMA Strategic Intelligence Solutions, Jordan} and {Public Opinion Research Institute, Kazakhstan} and {Central Asia Barometer, Kyrgyzstan} and {Statistics Lebanon Ltd.} and {University Of Macao, China} and {IPSOS Malaysia} and {Moreno \& Sotnikova Social Research And Consulting S.C., Mexico} and {IRL (Indochina Research Laos) Myanmar Limited, Myanmar} and {School Of People, Environment And Planning, Massey University, New Zealand} and {CID/Gallup, S.A. (Survey In Nicaragua)} and {TNS RMS Nigeria Limited / Kantar} and {Gallup Pakistan} and {Public Opinion Institute At Pontifical Catholic University Of Peru} and {Social Weather Stations, Philippines} and {Universidad Del Sagrado Coraz\'on, Puerto Rico} and {CESSI-Institute For Comparative Social Research (Survey In Russia)} and {Singidunum University Belgrade, Serbia} and {Gallup South Korea} and {Institute Of Sociology, Academia Sinica, Taipei, Taiwan ROC} and {Research Centre SHARQ /Oriens, Tajikistan} and {King Prajadhipok's Institute, Thailand} and {Applied Social Science Forum, Tunisia} and {Bahcesehir University, Turkey} and {University Of Chicago, NORC AmeriSpeak, USA} and {Indochina Research Ltd. Vietnam} and {Consumer Feedback, Zimbabwe \& TNS RMS Nigeria Limited/ Kantar} and {Social Monitoring Center} and {Info Sapiens Research Center} and {Ukrainian Center For European Policy, Ukraine.}},
  copyright = {Alle im GESIS DBK ver\"offentlichten Metadaten sind frei verf\"ugbar unter den Creative Commons CC0 1.0 Universal Public Domain Dedication. GESIS bittet jedoch darum, dass Sie alle Metadatenquellen anerkennen und sie nennen, etwa die Datengeber oder jeglichen Aggregator, inklusive GESIS selbst. F\"ur weitere Informationen siehe https://dbk.gesis.org/dbksearch/guidelines.asp?db=d, All metadata from GESIS DBK are available free of restriction under the Creative Commons CC0 1.0 Universal Public Domain Dedication. However, GESIS requests that you actively acknowledge and give attribution to all metadata sources, such as the data providers and any data aggregators, including GESIS. For further information see https://dbk.gesis.org/dbksearch/guidelines.asp},
  howpublished = {(ZA7505; Version 5.0.0) [Data set]. GESIS, Cologne.},
  langid = {ngerman}
}

@article{germann_dynamic_2016,
  title = {Dynamic scale validation reloaded: {{Assessing}} the psychometric properties of latent measures of ideology in {{VAA}} spatial maps},
  shorttitle = {Dynamic scale validation reloaded},
  author = {Germann, Micha and Mendez, Fernando},
  year = 2016,
  month = may,
  journal = {Quality \& Quantity},
  volume = {50},
  number = {3},
  pages = {981--1007},
  urldate = {2024-01-16},
  langid = {english}
}

@article{germann_spatial_2015,
  title = {Spatial maps in voting advice applications: {{The}} case for dynamic scale validation},
  shorttitle = {Spatial maps in voting advice applications},
  author = {Germann, Micha and Mendez, Fernando and Wheatley, Jonathan and Serd{\"u}lt, Uwe},
  year = 2015,
  month = apr,
  journal = {Acta Politica},
  volume = {50},
  number = {2},
  pages = {214--238},
  urldate = {2024-01-16},
  langid = {english}
}

@article{grung_missing_1998,
  title = {Missing values in principal component analysis},
  author = {Grung, Bj{\o}rn and Manne, Rolf},
  year = 1998,
  month = aug,
  journal = {Chemometrics and Intelligent Laboratory Systems},
  volume = {42},
  number = {1-2},
  pages = {125--139},
  publisher = {Elsevier BV},
  urldate = {2025-07-16},
  copyright = {https://www.elsevier.com/tdm/userlicense/1.0/},
  langid = {english}
}

@article{hinich_new_1981,
  title = {A {{New Approach}} to the {{Spatial Theory}} of {{Electoral Competition}}},
  author = {Hinich, Melvin J. and Pollard, Walker},
  year = 1981,
  month = may,
  journal = {American Journal of Political Science},
  volume = {25},
  number = {2},
  eprint = {2110856},
  eprinttype = {jstor},
  pages = {323},
  urldate = {2025-09-05},
  langid = {english}
}

@article{imai_fast_2016,
  title = {Fast {{Estimation}} of {{Ideal Points}} with {{Massive Data}}},
  author = {Imai, Kosuke and Lo, James and Olmsted, Jonathan},
  year = 2016,
  month = nov,
  journal = {American Political Science Review},
  volume = {110},
  number = {4},
  pages = {631--656},
  urldate = {2023-01-17},
  langid = {english}
}

@article{ioannidis_power_2025,
  title = {The power of alignment: how personalized information shapes voter decisions},
  shorttitle = {The power of alignment},
  author = {Ioannidis, Nikandros},
  year = 2025,
  month = jun,
  journal = {Journal of Information Technology \& Politics},
  pages = {1--22},
  urldate = {2025-06-17},
  langid = {english}
}

@article{jackman_multidimensional_2001,
  title = {Multidimensional {{Analysis}} of {{Roll Call Data}} via {{Bayesian Simulation}}: {{Identification}}, {{Estimation}}, {{Inference}}, and {{Model Checking}}},
  shorttitle = {Multidimensional {{Analysis}} of {{Roll Call Data}} via {{Bayesian Simulation}}},
  author = {Jackman, Simon},
  year = 2001,
  month = jan,
  journal = {Political Analysis},
  volume = {9},
  number = {3},
  pages = {227--241},
  urldate = {2023-09-07},
  langid = {english}
}

@misc{jackman_pscl_2024,
  title = {pscl: {{Classes}} and {{Methods}} for {{R}}. {{Developed}} in the {{Political Science Computational Laboratory}}},
  author = {Jackman, Simon},
  year = 2024,
  howpublished = {University of Sydney}
}

@article{jeon_mapping_2021,
  title = {Mapping {{Unobserved Item}}--{{Respondent Interactions}}: {{A Latent Space Item Response Model}} with {{Interaction Map}}},
  shorttitle = {Mapping {{Unobserved Item}}--{{Respondent Interactions}}},
  author = {Jeon, Minjeong and Jin, Ick Hoon and Schweinberger, Michael and Baugh, Samuel},
  year = 2021,
  month = jun,
  journal = {Psychometrika},
  volume = {86},
  number = {2},
  pages = {378--403},
  urldate = {2026-05-16},
  langid = {english}
}

@inproceedings{kingma_autoencoding_2014,
  title = {Auto-{{Encoding Variational Bayes}}},
  booktitle = {2nd {{Proceedings}} of the {{International Conference}} on {{Learning Representations}} ({{ICLR}}).},
  author = {Kingma, Diederik P. and Welling, Max},
  year = 2014,
  month = apr,
  eprint = {1312.6114},
  primaryclass = {cs, stat},
  urldate = {2024-01-15},
  archiveprefix = {arXiv},
  langid = {english}
}

@article{kobak_art_2019,
  title = {The art of using t-{{SNE}} for single-cell transcriptomics},
  author = {Kobak, Dmitry and Berens, Philipp},
  year = 2019,
  month = nov,
  journal = {Nature Communications},
  volume = {10},
  number = {1},
  pages = {5416},
  publisher = {Nature Publishing Group},
  urldate = {2026-05-18},
  copyright = {2019 The Author(s)},
  langid = {english}
}

@misc{lee_euclidean_2025,
  howpublished = {arXiv preprint arXiv:2512.11610},
  title = {Euclidean {{Ideal Point Estimation From Roll-Call Data}} via {{Distance-Based Bipartite Network Models}}},
  author = {Lee, Seungju and Kim, In Kyun and Park, Jong Hee and Jin, Ick Hoon},
  year = 2025,
  month = dec,
  number = {arXiv:2512.11610},
  eprint = {2512.11610},
  primaryclass = {stat},
  publisher = {arXiv},
  urldate = {2025-12-16},
  archiveprefix = {arXiv},
  langid = {english}
}

@article{leimgruber_comparing_2010,
  title = {Comparing {{Candidates}} and {{Citizens}} in the {{Ideological Space}}},
  author = {Leimgruber, Philipp and Hangartner, Dominik and Leemann, Lucas},
  year = 2010,
  month = sep,
  journal = {Swiss Political Science Review},
  volume = {16},
  number = {3},
  pages = {499--531},
  urldate = {2024-07-22},
  copyright = {http://onlinelibrary.wiley.com/termsAndConditions\#vor},
  langid = {english}
}

@article{leuthold_making_2007,
  title = {Making the {{Political Landscape Visible}}: {{Mapping}} and {{Analyzing Voting Patterns}} in an {{Ideological Space}}},
  shorttitle = {Making the {{Political Landscape Visible}}},
  author = {Leuthold, Heinrich and Hermann, Michael and Fabrikant, Sara I.},
  year = 2007,
  month = oct,
  journal = {Environment and Planning B: Planning and Design},
  volume = {34},
  number = {5},
  pages = {785--807},
  urldate = {2025-08-26},
  copyright = {https://journals.sagepub.com/page/policies/text-and-data-mining-license},
  langid = {english}
}

@article{mccoy_variational_2018,
  title = {Variational {{Autoencoders}} for {{Missing Data Imputation}} with {{Application}} to a {{Simulated Milling Circuit}}},
  author = {McCoy, John T. and Kroon, Steve and Auret, Lidia},
  year = 2018,
  month = aug,
  journal = {IFAC-PapersOnLine},
  volume = {51},
  number = {21},
  pages = {141--146},
  urldate = {2023-12-08},
  langid = {english}
}

@book{mccullagh_generalized_1989,
  title = {Generalized {{Linear Models}}},
  author = {McCullagh, P.},
  year = 1989,
  month = jan,
  edition = {2},
  publisher = {Routledge},
  address = {Boca Raton},
}

@article{mcinnes_umap_2018,
  title = {{{UMAP}}: {{Uniform Manifold Approximation}} and {{Projection}}},
  shorttitle = {{{UMAP}}},
  author = {McInnes, Leland and Healy, John and Saul, Nathaniel and Gro{\ss}berger, Lukas},
  year = 2018,
  month = sep,
  journal = {Journal of Open Source Software},
  volume = {3},
  number = {29},
  pages = {861},
  urldate = {2026-05-24},
  langid = {english}
}

@article{montgomery_computerized_2013,
  title = {Computerized {{Adaptive Testing}} for {{Public Opinion Surveys}}},
  author = {Montgomery, Jacob M. and Cutler, Josh},
  year = 2013,
  month = aug,
  journal = {Political Analysis},
  volume = {21},
  number = {2},
  pages = {172--192},
  urldate = {2023-06-26},
  langid = {english}
}

@article{otjes_spatial_2014,
  title = {Spatial models in voting advice applications},
  author = {Otjes, Simon and Louwerse, Tom},
  year = 2014,
  month = dec,
  journal = {Electoral Studies},
  volume = {36},
  pages = {263--271},
  urldate = {2023-06-26},
  langid = {english}
}

@article{petrik_core_2010,
  title = {Core {{Concept}} ``{{Political Compass}}''},
  author = {Petrik, Andreas},
  year = 2010,
  month = jan,
  journal = {Journal of Social Science Education},
  volume = {9},
  number = {4},
  pages = {45--62},
  langid = {english}
}

@misc{politools_daten_2019,
  title = {Daten zu den {{Schweizer Nationalrats-}} und {{St\"anderatswahlen}} 2019 der {{Online-Wahlhilfe Smartvote}}},
  author = {{Politools}},
  year = 2019,
}

@misc{politools_daten_2023,
  title = {Daten zu den {{Schweizer Nationalrats-}} und {{St\"anderatswahlen}} 2023 der {{Online-Wahlhilfe Smartvote}}},
  author = {{Politools}},
  year = 2023,
}

@article{poole_spatial_1985,
  title = {A {{Spatial Model}} for {{Legislative Roll Call Analysis}}},
  author = {Poole, Keith T. and Rosenthal, Howard},
  year = 1985,
  month = may,
  journal = {American Journal of Political Science},
  volume = {29},
  number = {2},
  pages = {357},
  urldate = {2023-01-03},
  langid = {english}
}

@article{potthoff_estimating_2018,
  title = {Estimating {{Ideal Points}} from {{Roll-Call Data}}: {{Explore Principal Components Analysis}}, {{Especially}} for {{More Than One Dimension}}?},
  shorttitle = {Estimating {{Ideal Points}} from {{Roll-Call Data}}},
  author = {Potthoff, Richard},
  year = 2018,
  month = jan,
  journal = {Social Sciences},
  volume = {7},
  number = {2},
  pages = {12},
  urldate = {2023-01-17},
  langid = {english}
}

@inproceedings{rahimi_random_2007,
  title = {Random {{Features}} for {{Large-Scale Kernel Machines}}},
  booktitle = {Advances in {{Neural Information Processing Systems}}},
  author = {Rahimi, Ali and Recht, Benjamin},
  year = 2007,
  publisher = {Curran Associates, Inc.},
  urldate = {2026-05-21}
}

@book{rubin_multiple_1987,
  title = {Multiple {{Imputation}} for {{Nonresponse}} in {{Surveys}}},
  author = {Rubin, Donald B.},
  year = 1987,
  publisher = {John Wiley \& Sons, Ltd},
  address = {New York, NY, USA},
  urldate = {2026-05-21},
  langid = {english}
}

@book{scholkopf_learning_2002,
  title = {Learning with kernels: support vector machines, regularization, optimization, and beyond},
  shorttitle = {Learning with kernels},
  author = {Sch{\"o}lkopf, Bernhard and Smola, Alexander J.},
  year = 2002,
  series = {Adaptive computation and machine learning},
  publisher = {MIT Press},
  address = {Cambridge, MA},
  langid = {english}
}

@article{sigfrid_irt_2024,
  title = {{{IRT}} for voting advice applications: a multi-dimensional test that is adaptive and interpretable},
  shorttitle = {{{IRT}} for voting advice applications},
  author = {Sigfrid, Karl},
  year = 2024,
  month = mar,
  journal = {Quality \& Quantity},
  volume = {58},
  pages = {4137--4156},
  urldate = {2024-03-06},
  langid = {english}
}

@article{small_polis_2021,
  title = {Polis: {{Scaling}} deliberation by mapping high dimensional opinion spaces},
  shorttitle = {Polis},
  author = {Small, Christopher and Bjorkegren, Michael and Erkkil{\"a}, Timo and Shaw, Lynette and Megill, Colin},
  year = 2021,
  month = jul,
  journal = {RECERCA. Revista de Pensament i An\`alisi},
  series = {Universitat {{Jaume I Servei}} de {{Comunicacio}} i {{Publicacions}}},
  volume = {26},
  number = {2},
  pages = {1--26},
  urldate = {2022-11-15},
  langid = {english}
}

@article{stokes_spatial_1963,
  title = {Spatial {{Models}} of {{Party Competition}}},
  author = {Stokes, Donald E.},
  year = 1963,
  month = jun,
  journal = {American Political Science Review},
  volume = {57},
  number = {2},
  pages = {368--377},
  urldate = {2025-09-03},
  copyright = {https://www.cambridge.org/core/terms},
  langid = {english}
}

@article{thomeczek_one_2025,
  title = {One model to rule them all? {{Choosing}} two-dimensional spaces for {{European}} political landscapes with {{VAA}} data},
  shorttitle = {One model to rule them all?},
  author = {Thomeczek, Jan Philipp and others},
  year = 2025,
  month = feb,
  journal = {Journal of Information Technology \& Politics},
  pages = {1--10},
  urldate = {2025-03-13},
  langid = {english}
}

@misc{vaeth_rational_2025,
  howpublished = {Working paper},
  title = {Rational {{Voter Learning}}, {{Issue Alignment}}, and {{Polarization}}},
  author = {Vaeth, Martin},
  year = 2025,
  langid = {english}
}

@article{vandermaaten_visualizing_2008,
  title = {Visualizing data using t-{{SNE}}},
  author = {{van der Maaten}, Laurens and Hinton, Geoffrey},
  year = 2008,
  journal = {Journal of machine learning research},
  volume = {9},
  number = {11},
}

@misc{verboom_visualizing_2025,
  howpublished = {IEEE VIS 2025},
  title = {Visualizing {{Opinion Space}} in {{Voting Advice Applications}}: {{A User Study}}},
  author = {Verboom, Damion E and Mchedlidze, Tamara and Oral, Basak and Dimara, Evanthia and Rebelo, Daniela Peres and Kamoen, Naomi and Bearfield, Cindy Xiong},
  year = 2025,
  langid = {english}
}
\appendix
\clearpage
\section{Model Details}
\label{app:model-details}

\IXPLORE---short for \emph{Iterative X-Posterior LOgistic Regression Embedding}---exposes a small set of hyperparameters that control the prior, the optimization loop, and the per-item logistic regression.
Their operating values are listed in \cref{tab:hparams}, alongside which sub-paragraph of the configuration analysis (\cref{sec:results-configuration}) varies each one.
This section documents the design choices that affect \IXPLORE's behavior but are not central to the main text.

\begin{table}[ht]
\centering
\caption{\IXPLORE hyperparameters. The defaults are used in the baseline comparison (\cref{sec:results-baseline}), while the configuration analysis (\cref{sec:results-configuration}) varies parameters as described.}
\label{tab:hparams}
\small
\begin{tabular}{llll}
\toprule
Hyperparameter & Symbol & Default & Varied in \\
\midrule
Prior variance        & $\tau^2$            & $0.25$              & \cref{sec:results-prior} \\
Iterations            & $T$                   & $10$                 & \cref{sec:results-iteration} \\
Initialization        & --                    & PCA                 & \cref{sec:results-iteration} \\
Feature transform     & $\varphi$             & identity  & \cref{sec:results-features} \\
Confidence weights    & $W_{nk}$              & uniform  & \cref{sec:results-weights} \\
Weight scaling        & --                    & \texttt{False}      & \cref{sec:results-weights} \\
Grid resolution       & $R$                   & $200$               & $100$ in \cref{sec:results-configuration}, \cref{app:weights} \\
Item L2 penalty       & $\lambda$             & $10^{-8}$           & fixed \\
Grid half-extent      & $L$                   & $1.0$               & fixed \\
\bottomrule
\end{tabular}
\end{table}

\subsection{Gaussian Prior and Log-Barrier}
\label{app:prior-shape}

\begin{figure}[ht]
    \centering
    \includegraphics{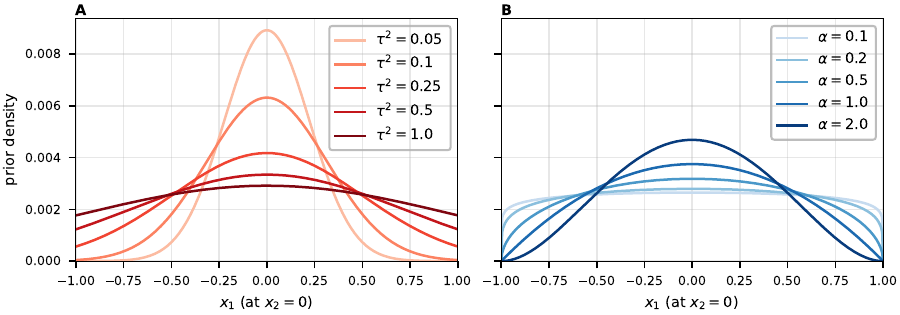}
    \caption{One-dimensional slice of the two prior families at $x_2 = 0$ over the grid $[-L, L]^2$ with $L = 1$. \abc{A}~The Gaussian prior concentrates around the origin as $\tau^2$ shrinks. \abc{B}~The log-barrier prior is flat in the interior and vanishes at the grid boundary for larger $\alpha$. Both priors are symmetric in $x_1$ and $x_2$.}
    \label{fig:prior_families}
\end{figure}

\IXPLORE supports two prior families over the latent position, shown in \cref{fig:prior_families}.
The Gaussian prior with variance $\tau^2$ is centrally peaked.
The log-barrier prior with strength $\alpha$ is flat in the interior and vanishes at the grid boundary.
We compare both on \smartvote 2023 under the default configuration ($T = 5$, candidates as training users), sweeping the Gaussian variance $\tau^2$ and the log-barrier strength $\alpha$, and additionally embedding voters with between 5 and 75 observed answers.

The reconstruction error is flat across every prior and strength (MAE in 0.1814--0.1831).
The boundary fraction varies from 0\% to 38\%, and collapses to 0\% under either a strong Gaussian ($\tau^2 \leq 0.1$) or a strong log-barrier ($\alpha \geq 2$).
At a matched boundary fraction of 0\%, the log-barrier retains more embedding spread than the Gaussian (spread 0.43 at $\alpha = 5$ versus 0.37 at $\tau^2 = 0.05$).
For voters embedded from 5 to 75 observed answers, the imputation MAE varies by at most 0.01 across the Gaussian, log-barrier, and uniform-prior configurations.

\subsection{Point Estimate from the Grid Posterior}
\label{app:point-estimate}

To summarize each user's grid posterior with a single 2D position $\tilde{x}_n$, three options are natural.
The \emph{posterior mean} is the probability-weighted average grid position of \cref{eq:discrete-posterior}
\[
    \tilde{x}_n \;=\; \sum_{g \in X_{\text{grid}}} g \, p(x_n = g \mid Y_n^{\text{obs}}, \theta).
\]
The \emph{MAP} is the single highest-posterior grid point
\[
    \hat{x}_n \;=\; \arg\max_{g \in X_{\text{grid}}} p(x_n = g \mid Y_n^{\text{obs}}, \theta).
\]
\emph{Full marginalization} avoids collapsing the posterior at all: predictions for new items integrate the response function against the posterior, the discrete counterpart of \cref{eq:posterior-predictive-pk},
\[
    \hat{Y}_{nk} \;=\; \sum_{g \in X_{\text{grid}}} y_k(g)\, p(x_n = g \mid Y_n^{\text{obs}}, \theta).
\]
We compare the three on \smartvote 2023 ($N{=}1029$, $K{=}75$) at the default configuration ($\tau^2 = 0.25$, $T = 10$).

\paragraph{Predictive accuracy.}
The posterior mean yields the lowest held-out MAE at every observation level (\cref{tab:point-estimate-mae}).
Its gap to the MAP is largest with few observed answers ($\Delta\text{MAE} = 0.0058$ at $n_{\text{obs}}{=}5$) and narrows to 0.0014 at $n_{\text{obs}}{=}70$.
Full marginalization is consistently the worst of the three: the smoothing it provides does not compensate for predicting a probability that need not match any single grid point.

\paragraph{Latent geometry.}
The MAP estimator is constrained to the $G = R^2$ grid coordinates, which produces lattice artifacts in the scatter plot at any practical resolution and makes it pile up on the boundary as the prior weakens: the fraction of embeddings within 5\% of the grid edge grows from 0.000 at $\tau^2 = 0.05$ to 0.149 at $\tau^2 = 10^6$ (\cref{tab:point-estimate-boundary}).
The posterior mean, pulled inward by the probability-weighted average, stays at or below 0.007 across the whole range while the reconstruction quality is essentially identical between the two estimators: MAE stays near 0.181 and ACC at 0.835--0.836 for every $\tau^2$.

\begin{table}[ht]
\centering
\caption{Held-out MAE on \smartvote 2023 by point estimator and number of observed answers, at the default configuration ($\tau^2 = 0.25$, $T = 10$). Best per row in \textbf{bold}.}
\label{tab:point-estimate-mae}
\begin{tabular}{rccc}
\toprule
$n_{\text{obs}}$ & MAP & Posterior mean & Full posterior \\
\midrule
5 & 0.2313 & \textbf{0.2255} & 0.2439 \\
10 & 0.2111 & \textbf{0.2067} & 0.2190 \\
15 & 0.2023 & \textbf{0.1983} & 0.2078 \\
30 & 0.1934 & \textbf{0.1905} & 0.1961 \\
45 & 0.1915 & \textbf{0.1894} & 0.1933 \\
60 & 0.1894 & \textbf{0.1878} & 0.1907 \\
70 & 0.1924 & \textbf{0.1910} & 0.1935 \\
\bottomrule
\end{tabular}

\end{table}

\begin{table}[ht]
\centering
\caption{Boundary fraction (share of embeddings within 5\% of the grid edge) and train reconstruction MAE / accuracy on \smartvote 2023, per prior variance with a model retrained at each $\tau^2$ ($T = 10$), for the MAP and posterior-mean estimators.}
\label{tab:point-estimate-boundary}
\begin{tabular}{rcccccc}
\toprule
 & \multicolumn{2}{c}{Boundary frac.} & \multicolumn{2}{c}{Recon. MAE} & \multicolumn{2}{c}{Recon. Acc.} \\
\cmidrule(lr){2-3}\cmidrule(lr){4-5}\cmidrule(lr){6-7}
$\tau^2$ & MAP & Mean & MAP & Mean & MAP & Mean \\
\midrule
0.05 & 0.000 & 0.000 & 0.1824 & \textbf{0.1804} & \textbf{0.836} & 0.836 \\
0.1 & 0.000 & 0.000 & 0.1823 & \textbf{0.1804} & 0.836 & \textbf{0.836} \\
0.25 & 0.006 & \textbf{0.000} & 0.1820 & \textbf{0.1805} & 0.836 & \textbf{0.836} \\
0.5 & 0.045 & \textbf{0.000} & 0.1815 & \textbf{0.1806} & 0.836 & \textbf{0.836} \\
1 & 0.088 & \textbf{0.000} & 0.1811 & \textbf{0.1807} & \textbf{0.836} & 0.836 \\
$10^{6}$ & 0.149 & \textbf{0.007} & \textbf{0.1807} & 0.1808 & \textbf{0.835} & 0.835 \\
\bottomrule
\end{tabular}

\end{table}

The posterior mean is therefore set as the default as it achieves the lowest held-out MAE at every observation level, matches reconstruction quality with the MAP, and raises no boundary artifacts.

\subsubsection{Vectorized Grid Evaluation}
\label{app:vectorization}

The dominant per-iteration cost is evaluating the Bernoulli log-likelihood at the $G$ grid points for each of $N$ users against $K$ items, naively $\mathcal{O}(NKG)$.
Defining the masked answer matrices $\tilde{Y}_{nk} = M_{nk} Y_{nk}$ and $\tilde{Z}_{nk} = M_{nk}(1 - Y_{nk})$, and the per-grid-cell log-prediction tables $\mathrm{LP}_{gk} = \log y_k(g)$ and $\mathrm{LN}_{gk} = \log(1 - y_k(g))$, the full $(N, G)$ matrix of user log-likelihoods is
\begin{equation}
    L \;=\; \tilde{Y}\, \mathrm{LP}^\top \;+\; \tilde{Z}\, \mathrm{LN}^\top,
    \label{eq:vectorized-loglik}
\end{equation}
two BLAS matrix products of shape $(N, K) \times (K, G)$, which replaces the triple loop with two dense matrix multiplications.

\subsubsection{Sampling Strategies}
\label{app:sampling}

For applications that require synthetic answer vectors rather than point predictions---e.g., data augmentation or active-learning ensembles---\IXPLORE supports two sampling strategies in addition to a uniform-random baseline, following the principle of multiple imputation~\citep{rubin_multiple_1987}.
\emph{Posterior sampling} draws latent positions from the grid posterior $p(x_n = g \mid Y_n^{\text{obs}}, \theta)$ and predicts continuous answer probabilities from each sampled position via the fitted item models, so the output is continuous in $[0,1]$ and the sample variance reflects the posterior spread.
\emph{Rasch sampling} first imputes all missing answers via posterior marginalization, then, for each item, evaluates normal category-response functions centered at each discrete answer option (e.g.\ $\{0, 0.25, 0.5, 0.75, 1\}$ for a 5-point scale) and samples a category by the Gumbel-max trick, producing discrete output whose variance is set by the CRF bandwidth.

We compare the two on \smartvote 2023 under the same protocol as the point-estimate study (\cref{app:point-estimate}): for each candidate and each $n_{\text{obs}}$, a random subset of $n_{\text{obs}}$ answers is revealed; we measure (i) the MAE between the per-item sample mean and the true answer on the remaining items, and (ii) the per-item sample standard deviation on those items.

\begin{figure}[ht]
    \centering
    \includegraphics{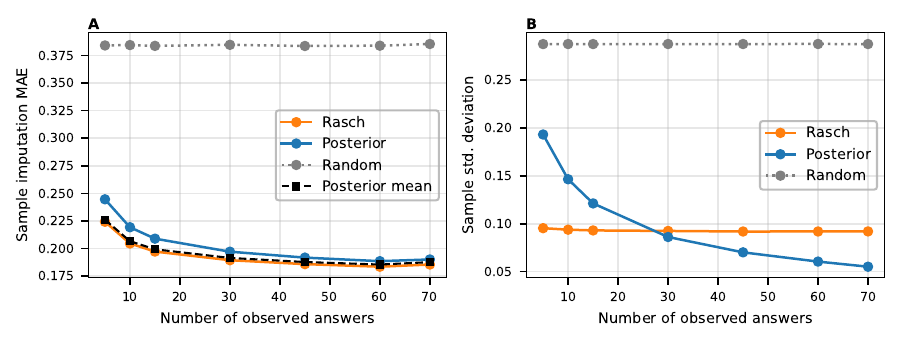}
    \caption{Sampling strategies on \smartvote 2023 as a function of the number of observed answers. \abc{A} The MAE is computed between each method's per-item sample mean and the true answer on the unobserved items, for Rasch, posterior, and uniform-random sampling, with the deterministic posterior-mean prediction for reference. \abc{B} The mean per-item sample standard deviation on the unobserved items is reported; higher values indicate more diverse samples.}
    \label{fig:sampling_strategies}
\end{figure}

Both strategies track the deterministic posterior-mean prediction closely on MAE at every sparsity level, while the uniform-random baseline has substantially higher MAE.
They differ in sample diversity: posterior samples inherit their spread from the posterior, so variance is high when the user has answered few questions and shrinks as the posterior concentrates; Rasch samples have variance fixed by the CRF bandwidth, which makes them less diverse for sparse users but more interpretable as plausible discrete answers.

Since both strategies preserve predictive accuracy while adding well-calibrated diversity, neither is set as a default.
The choice depends on the downstream use: posterior sampling for continuous augmentation that reflects posterior uncertainty, Rasch sampling for discrete pseudo-answers in the model's native answer space.

\subsubsection{Posterior Convergence on Test Users}
\label{app:posterior-convergence}

\begin{figure}[t]
\centering
\includegraphics{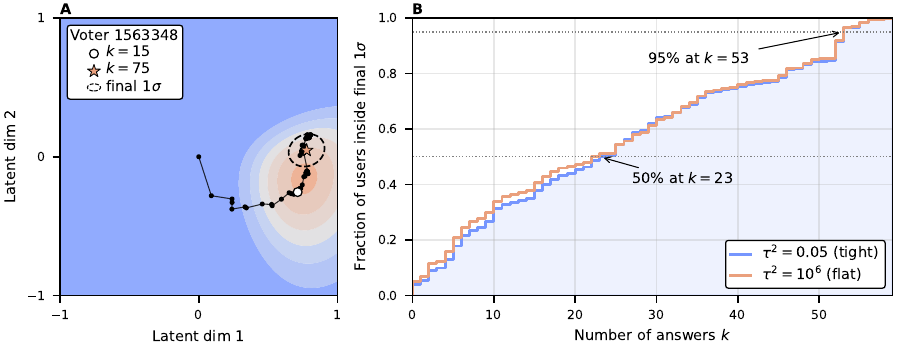}
\caption{Posterior trajectory and convergence on \smartvote 2023.
\abc{A} The test user whose final posterior mean lies furthest from the origin is shown, fitted at $\tau^2 = 0.25$. The contour is the posterior density after the user has answered $k = 15$ of $K = 75$ items; black dots and the connecting line trace the running posterior mean at every $k$ in answer order; the dashed ellipse is the $1\sigma$ contour of the moment-matched Gaussian fit to the final posterior.
\abc{B} The CDF of the smallest $k$ at which the running posterior mean enters the final $1\sigma$ ellipse is shown, evaluated over all 1,121 test users, for the tightest ($\tau^2 = 0.05$) and weakest ($\tau^2 = 10^6$) priors in our sweep. Dotted lines mark the 50\% and 95\% crossings.}
\label{fig:posterior-trajectory}
\end{figure}

We track at which item $k$ each test user's running posterior mean first enters the $1\sigma$ ellipse of their final posterior, as a measure of how many answers suffice to localize the user up to posterior uncertainty (\cref{fig:posterior-trajectory}).
Half of the held-out voters are inside the final $1\sigma$ ellipse after $k=23$ items and 95\% by $k=53$---roughly a third of the questionnaire suffices for the median user.
This rate is essentially insensitive to the prior: shrinking $\tau^2$ from $10^6$ (effectively flat) to 0.05 leaves the CDF unchanged within one item across the whole range.

\clearpage
\section{Reference Algorithms}
\label{app:reference-algorithms}

All baselines embed into a 2-dimensional latent space to be visualizable as a political spectrum (\emIRT excepted, see below).
When requiring binary input, each algorithm binarizes responses at threshold 0.5 before fitting.
Neutral responses at exactly 0.5 are treated as missing.
When an algorithm does not have native inference, a 2D embedding is fitted, and a per-item binary logistic decoder $\hat{Y}_{nk} = \sigma(x_n^\top \beta_k + \alpha_k)$ predicts the answers from the embedded position.
Where necessary, sparse held-out (test) users are imputed by the train-column means.

\paragraph{PCA.}
The pipeline of~\citet{potthoff_estimating_2018} represents each user by a $d$-dimensional PCA score.
\emph{Linear PCA} skips the logistic decoder and instead reconstructs $\hat{Y}_{nk} = (x_n^\top v_k + \mu_k)$ clipped to $[0, 1]$, where $(v_k, \mu_k)$ are the rank-$d$ PCA loadings and column mean, with missing training entries filled by column means before a single PCA fit.
\emph{Logistic PCA} uses the shared logistic decoder, with missing entries handled by the iterative-PCA imputation of~\citet{grung_missing_1998} (column-mean initialization, then alternating rank-$d$ reconstruction until convergence).
For the main analysis in \cref{sec:results-baseline}, we choose $d=2$ to match \IXPLORE's latent space; for the dimensionality study of \cref{sec:results-dimensions}, we sweep $d \in \{1, ..., 74\}$.

\paragraph{Kernel PCA.}
The embedding stage is an RBF-kernel PCA~\citep{scholkopf_learning_2002}, the nonlinear counterpart of the linear PCA above, with missing entries mean-imputed before the kernel is computed.
The kernel bandwidth $\gamma$ is the only hyper-parameter; its useful scale is dataset-dependent, so it is selected on the training reactions alone: a random 15\% of the observed training entries is held out, kernel PCA and the decoder are refitted for each candidate $\gamma$ on the remaining entries, and the $\gamma$ with the lowest mean absolute error on the held-out entries is kept.
The candidate grid is expressed as the multipliers $\{0.1, 0.25, 0.5, 1, 2, 4\}$ of the RBF median heuristic ($\gamma = 1 / \mathrm{median}$ pairwise squared distance) so that it adapts to each dataset's scale.
Held-out users are projected with \textsf{scikit-learn}'s kernel-PCA out-of-sample transform against the fitted training points.

\paragraph{t-SNE.}
The embedding stage is \tSNE (perplexity 30), using the \textsf{openTSNE} implementation so that held-out users can be projected into the fitted space with its out-of-sample transform.

\paragraph{UMAP.}
The embedding stage is \UMAP (15 neighbors, minimum distance 0.1); a fixed random state is used for reproducibility, which forces single-threaded execution.
Held-out users are projected with \UMAP's out-of-sample transform.

\paragraph{Variational Auto-Encoder.}
VAEs learn the embedding and the decoder jointly as one generative model.
We use the variational auto-encoder~\citep{kingma_autoencoding_2014} in the configuration of the VAE baseline of~\citet{bachmann_fast_2024}.
The encoder is a single 64-unit ReLU hidden layer followed by two heads producing $\mu(y)$ and $\log \sigma^2(y)$ for a 2D Gaussian posterior over the latent $z$.
We compare two decoders: a \emph{logistic} decoder $\sigma(W z + b)$ and a \emph{2-layer} decoder with a 64-unit ReLU hidden layer and a sigmoid output.
Training minimizes a mask-renormalized reconstruction loss plus a KL term,
\begin{equation}
    \mathcal{L} \;=\; \frac{\sum_{(n,k) \in \mathcal{O}} (\hat{Y}_{nk} - Y_{nk})^2}{|\mathcal{O}|/(NK)} \;+\; \beta \cdot \mathrm{KL}\left(q(z \mid Y) \,\|\, \mathcal{N}(0, I)\right),
\end{equation}
with $\beta = 1$ for the 2-layer decoder and $\beta = 0$ for the logistic decoder, mean-imputation of missing entries at the encoder input, and a binary observation mask on the reconstruction term.
We optimize with Adam ($\text{lr} = 10^{-3}$) and early-stop on a 10\% held-out validation split with patience 300.
Predictions for any cell $(n, k)$ are obtained by passing the user's mean-imputed answer vector through the encoder, taking the posterior mean as the latent code, and reading off the decoder output at index $k$.

\paragraph{IDEAL.}
We use the Bayesian item-response model of~\citet{clinton_statistical_2004} and \citet{jackman_multidimensional_2001} with the probit likelihood
\begin{equation}
    \Pr(Y_{nk} = 1 \mid x_n, \beta_k, \alpha_k) \;=\; \Phi\bigl(x_n^\top \beta_k - \alpha_k\bigr),
\end{equation}
as implemented in the \textsf{pscl} R package~\citep{jackman_pscl_2024}.
Item parameters are fitted on the training matrix by MCMC (10,000 iterations, 5,000 burn-in, thinning 100); the wrapper retains the posterior-mean discriminations $\beta_k$ and difficulties $\alpha_k$.
To embed a held-out user, we run a second MCMC chain with item parameters held fixed via spike priors (variance $10^6$ on $\beta$), so only the new user's ideal point is updated.
Predictions for any cell $(n, k)$ are $\hat{Y}_{nk} = \Phi(x_n^\top \beta_k - \alpha_k)$ at the posterior-mean estimate of $x_n$.

\paragraph{emIRT.}
We use the fast expectation-maximization estimator of~\citet{imai_fast_2016} for the binary ideal point model, as implemented by the \texttt{binIRT} routine of the \textsf{\emIRT} R package.
This is the EM counterpart of \IDEAL: it maximizes the same probit likelihood
\begin{equation}
    \Pr(Y_{nk} = 1 \mid x_n, \beta_k, \alpha_k) \;=\; \Phi\bigl(\beta_k\, x_n - \alpha_k\bigr),
\end{equation}
but recovers point estimates by EM rather than MCMC, so it is included as a fast-estimation reference on the same 0.5-binarized data as \IDEAL.
The estimator is one-dimensional only; we expose the single estimated ideal point as the first embedding coordinate and fix the second at 0, so \emIRT appears as a horizontal line in the two-dimensional embedding plots.
Item parameters $(\beta_k, \alpha_k)$ are fitted on the training matrix by EM (convergence threshold $10^{-6}$).
To embed a held-out user, the item parameters are held fixed and the user's ideal point is obtained by maximizing their binary log-likelihood, the same one-dimensional counterpart of the procedure used for the other reference algorithms.
Predictions for any cell $(n, k)$ are $\hat{Y}_{nk} = \Phi(\beta_k\, x_n - \alpha_k)$ at the estimated $x_n$.

\paragraph{LSIRM.}
We use the latent-space item-response model of~\citet{jeon_mapping_2021}, in the metric-space formulation of~\citet{lee_euclidean_2025}, as implemented by the \texttt{lsirm2pl} routine of the \textsf{lsirm12pl} R package.
Each user $n$ and item $k$ receive a 2D latent position $z_n$ and $w_k$, and the response probability decreases with their Euclidean distance,
\begin{equation}
    \mathrm{logit}\,\Pr(Y_{nk} = 1 \mid z_n, w_k, \theta_n, \alpha_k, \beta_k) \;=\; \theta_n\, \alpha_k + \beta_k - \gamma\, \lVert z_n - w_k \rVert,
\end{equation}
where $\theta_n$ is a per-user random effect, $\alpha_k$ and $\beta_k$ are item discrimination and difficulty, and $\gamma$ scales the distance term.
The user positions $z_n$ are the embedding.
Item parameters and positions are fitted on the training matrix by MCMC (5,000 iterations, 1,000 burn-in, thinning 5, missing entries handled under the missing-at-random model); the wrapper retains the posterior-mean estimates.
To embed a held-out user, the item parameters $(w_k, \alpha_k, \beta_k, \gamma)$ are held fixed and the user's $(z_n, \theta_n)$ is the maximum-a-posteriori estimate under the same prior the training MCMC uses ($z_n \sim \mathcal{N}(0, I)$, $\theta_n \sim \mathcal{N}(0, 1)$), so out-of-sample positions stay in the same regularized regime as the training posterior means rather than diverging under an unconstrained fit.
Predictions for any cell $(n, k)$ are $\hat{Y}_{nk} = \sigma\bigl(\theta_n \alpha_k + \beta_k - \gamma \lVert z_n - w_k \rVert\bigr)$ at the estimated $(z_n, \theta_n)$.

\clearpage
\section{Additional Tables and Figures}
\label{app:results}

\begin{table}[ht]
\centering
\caption{Performance comparison on Smartvote (2019). Each cell is the mean over all sparsity levels. Reference algorithms are listed in the top block; \IXPLORE is run both with the continuous input and the binarized input$^*$. The best-performing algorithm is shown in \textbf{bold}, the runner-up in \textit{italics}.}
\label{tab:baseline_smartvote_2019}
\small
\begin{tabular}{lcccccccc}
\toprule
 & \multicolumn{2}{c}{Train Rec.} & \multicolumn{2}{c}{Train Imp.} & \multicolumn{2}{c}{Test Rec.} & \multicolumn{2}{c}{Test Imp.} \\
\cmidrule(lr){2-3}\cmidrule(lr){4-5}\cmidrule(lr){6-7}\cmidrule(lr){8-9}
Algorithm & MAE & ACC & MAE & ACC & MAE & ACC & MAE & ACC \\
\midrule
Linear PCA & 0.239 & 80.0\% & 0.265 & 76.6\% & 0.234 & 81.1\% & 0.260 & 78.1\% \\
Logistic PCA & 0.176 & 85.0\% & 0.207 & \textbf{80.5\%} & 0.212 & 80.8\% & 0.236 & 77.8\% \\
Kernel PCA & 0.199 & 84.0\% & 0.227 & 79.6\% & 0.232 & 79.4\% & 0.256 & 76.2\% \\
t-SNE & 0.213 & 80.8\% & 0.239 & 76.5\% & 0.221 & 79.7\% & 0.242 & 76.6\% \\
UMAP & 0.207 & 81.1\% & 0.228 & 77.6\% & 0.228 & 78.4\% & 0.245 & 76.1\% \\
VAE (2-layer) & 0.192 & 83.9\% & 0.213 & \textit{80.4\%} & 0.227 & 80.1\% & 0.250 & 77.2\% \\
VAE (logistic) & \textit{0.174} & 85.1\% & \textit{0.206} & 80.0\% & 0.217 & 81.1\% & 0.244 & 77.5\% \\
IDEAL & 0.183 & 84.5\% & 0.212 & 79.7\% & 0.178 & 84.8\% & \textit{0.200} & 81.2\% \\
emIRT & 0.206 & 81.5\% & 0.224 & 78.5\% & 0.206 & 81.8\% & 0.236 & 78.0\% \\
LSIRM & 0.199 & 83.3\% & 0.227 & 78.4\% & 0.178 & \textbf{86.5\%} & 0.214 & 80.4\% \\
\midrule
IXPLORE & \textbf{0.170} & \textit{85.1\%} & \textbf{0.201} & 80.3\% & \textbf{0.169} & 85.4\% & \textbf{0.193} & \textbf{82.0\%} \\
IXPLORE$^*$ & 0.179 & \textbf{86.0\%} & 0.219 & 79.7\% & \textit{0.173} & \textit{86.0\%} & 0.200 & \textit{81.6\%} \\
\bottomrule
\end{tabular}
\end{table}

\begin{table}[ht]
\centering
\caption{Performance comparison on Polis (vTaiwan). Each cell is the mean over all sparsity levels. Reference algorithms are listed in the top block; \IXPLORE is run both with the continuous input and the binarized input$^*$. The best-performing algorithm is shown in \textbf{bold}, the runner-up in \textit{italics}.}
\label{tab:baseline_polis}
\small
\begin{tabular}{lcccccccc}
\toprule
 & \multicolumn{2}{c}{Train Rec.} & \multicolumn{2}{c}{Train Imp.} & \multicolumn{2}{c}{Test Rec.} & \multicolumn{2}{c}{Test Imp.} \\
\cmidrule(lr){2-3}\cmidrule(lr){4-5}\cmidrule(lr){6-7}\cmidrule(lr){8-9}
Algorithm & MAE & ACC & MAE & ACC & MAE & ACC & MAE & ACC \\
\midrule
Linear PCA & 0.281 & 82.4\% & 0.307 & 78.5\% & 0.282 & 81.4\% & 0.300 & 79.3\% \\
Logistic PCA & 0.178 & 88.9\% & \textbf{0.240} & \textit{80.9\%} & 0.253 & 81.2\% & 0.271 & 79.1\% \\
Kernel PCA & 0.259 & 85.2\% & 0.297 & 79.2\% & 0.274 & 82.0\% & 0.289 & 79.9\% \\
t-SNE & 0.272 & 82.2\% & 0.313 & 75.1\% & 0.271 & 82.6\% & 0.290 & 79.2\% \\
UMAP & 0.281 & 82.0\% & 0.315 & 74.9\% & 0.324 & 75.5\% & 0.335 & 73.1\% \\
VAE (2-layer) & 0.210 & 87.5\% & 0.256 & \textbf{81.0\%} & 0.245 & 84.3\% & 0.271 & \textit{80.9\%} \\
VAE (logistic) & 0.194 & 88.5\% & 0.253 & 80.2\% & 0.232 & 85.6\% & 0.262 & \textbf{81.4\%} \\
IDEAL & 0.199 & 87.7\% & 0.258 & 79.6\% & 0.228 & 86.1\% & 0.266 & 80.2\% \\
emIRT & 0.217 & 86.5\% & 0.258 & 80.5\% & 0.228 & 84.1\% & 0.288 & 76.2\% \\
LSIRM & 0.238 & 84.4\% & 0.276 & 78.8\% & 0.244 & 86.9\% & 0.278 & 80.5\% \\
\midrule
IXPLORE & \textit{0.164} & \textit{89.8\%} & 0.256 & 78.8\% & \textit{0.190} & \textit{87.8\%} & \textbf{0.251} & 79.3\% \\
IXPLORE$^*$ & \textbf{0.159} & \textbf{90.8\%} & \textit{0.252} & 79.1\% & \textbf{0.189} & \textbf{88.4\%} & \textit{0.254} & 79.2\% \\
\bottomrule
\end{tabular}
\end{table}

\begin{table}[ht]
\centering
\caption{Performance comparison on Voteview (S117). Each cell is the mean over all sparsity levels. Reference algorithms are listed in the top block; \IXPLORE is run both with the continuous input and the binarized input$^*$. The best-performing algorithm is shown in \textbf{bold}, the runner-up in \textit{italics}.}
\label{tab:baseline_voteview}
\small
\begin{tabular}{lcccccccc}
\toprule
 & \multicolumn{2}{c}{Train Rec.} & \multicolumn{2}{c}{Train Imp.} & \multicolumn{2}{c}{Test Rec.} & \multicolumn{2}{c}{Test Imp.} \\
\cmidrule(lr){2-3}\cmidrule(lr){4-5}\cmidrule(lr){6-7}\cmidrule(lr){8-9}
Algorithm & MAE & ACC & MAE & ACC & MAE & ACC & MAE & ACC \\
\midrule
Linear PCA & 0.231 & 90.6\% & 0.320 & 82.9\% & 0.267 & 90.9\% & 0.318 & 89.9\% \\
Logistic PCA & 0.053 & 96.8\% & 0.082 & \textit{93.6\%} & 0.190 & 84.7\% & 0.226 & 81.8\% \\
Kernel PCA & 0.121 & 95.2\% & 0.169 & 90.9\% & 0.192 & 85.1\% & 0.222 & 81.7\% \\
t-SNE & 0.124 & 95.1\% & 0.183 & 89.0\% & 0.243 & 77.4\% & 0.297 & 71.5\% \\
UMAP & 0.143 & 93.6\% & 0.208 & 87.0\% & 0.124 & 89.7\% & 0.140 & 88.3\% \\
VAE (2-layer) & 0.061 & 96.9\% & 0.092 & 93.4\% & 0.189 & 86.8\% & 0.224 & 84.0\% \\
VAE (logistic) & 0.092 & 93.8\% & 0.108 & 91.8\% & 0.332 & 68.5\% & 0.389 & 62.6\% \\
IDEAL & 0.045 & 96.3\% & \textbf{0.072} & 93.2\% & 0.078 & \textbf{94.1\%} & \textbf{0.083} & 93.3\% \\
emIRT & 0.074 & 95.9\% & 0.091 & \textbf{94.0\%} & 0.099 & 93.8\% & 0.100 & \textbf{93.5\%} \\
LSIRM & 0.099 & 94.7\% & 0.122 & 92.7\% & 0.098 & 93.9\% & 0.104 & \textit{93.3\%} \\
\midrule
IXPLORE & \textbf{0.032} & \textbf{97.7\%} & \textit{0.080} & 92.7\% & \textbf{0.076} & \textit{94.0\%} & \textit{0.083} & 93.1\% \\
IXPLORE$^*$ & \textit{0.032} & \textit{97.7\%} & 0.080 & 92.7\% & \textit{0.076} & 94.0\% & 0.083 & 93.1\% \\
\bottomrule
\end{tabular}
\end{table}

\begin{table}[ht]
\centering
\caption{Performance comparison on European Values Study. Each cell is the mean over all sparsity levels. Reference algorithms are listed in the top block; \IXPLORE is run both with the continuous input and the binarized input$^*$. The best-performing algorithm is shown in \textbf{bold}, the runner-up in \textit{italics}.}
\label{tab:baseline_evs}
\small
\begin{tabular}{lcccccccc}
\toprule
 & \multicolumn{2}{c}{Train Rec.} & \multicolumn{2}{c}{Train Imp.} & \multicolumn{2}{c}{Test Rec.} & \multicolumn{2}{c}{Test Imp.} \\
\cmidrule(lr){2-3}\cmidrule(lr){4-5}\cmidrule(lr){6-7}\cmidrule(lr){8-9}
Algorithm & MAE & ACC & MAE & ACC & MAE & ACC & MAE & ACC \\
\midrule
Linear PCA & 0.207 & 75.6\% & 0.230 & \textit{71.3\%} & 0.207 & 75.9\% & 0.229 & 71.6\% \\
Logistic PCA & \textit{0.182} & \textit{79.6\%} & 0.240 & 71.2\% & \textit{0.199} & 76.2\% & \textit{0.225} & \textbf{71.9\%} \\
Kernel PCA & 0.195 & 76.5\% & \textbf{0.226} & \textbf{71.5\%} & 0.202 & 76.1\% & 0.227 & 71.7\% \\
t-SNE & 0.204 & 74.8\% & 0.235 & 70.1\% & 0.209 & 74.1\% & 0.231 & 70.7\% \\
UMAP & 0.208 & 74.3\% & 0.236 & 69.6\% & 0.205 & 74.6\% & 0.229 & 70.9\% \\
VAE (2-layer) & 0.220 & 74.8\% & 0.236 & 70.6\% & 0.243 & 70.7\% & 0.246 & 69.3\% \\
VAE (logistic) & -- & -- & -- & -- & -- & -- & -- & -- \\
IDEAL & -- & -- & -- & -- & -- & -- & -- & -- \\
emIRT & 0.213 & 77.0\% & 0.254 & 69.8\% & 0.260 & 75.2\% & 0.349 & 63.1\% \\
LSIRM & -- & -- & -- & -- & -- & -- & -- & -- \\
\midrule
IXPLORE & \textbf{0.173} & 78.8\% & \textit{0.230} & 69.9\% & \textbf{0.180} & \textit{77.8\%} & \textbf{0.221} & 71.3\% \\
IXPLORE$^*$ & 0.204 & \textbf{82.2\%} & 0.270 & 70.1\% & 0.200 & \textbf{80.0\%} & 0.238 & \textit{71.7\%} \\
\bottomrule
\end{tabular}
\end{table}

\begin{figure}[ht]
    \centering
    \includegraphics{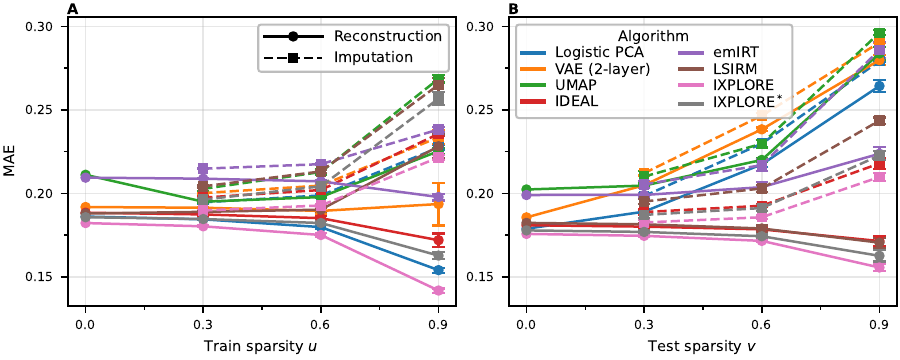}
    \caption{Reconstruction (solid, circles) and imputation (dashed, squares) MAE on \smartvote 2019 as a function of train sparsity $u$ and test sparsity $v$. Error bars show the standard deviation across seeds.}
    \label{fig:baseline_train_test_smartvote_2019}
\end{figure}

\begin{figure}[ht]
    \centering
    \includegraphics{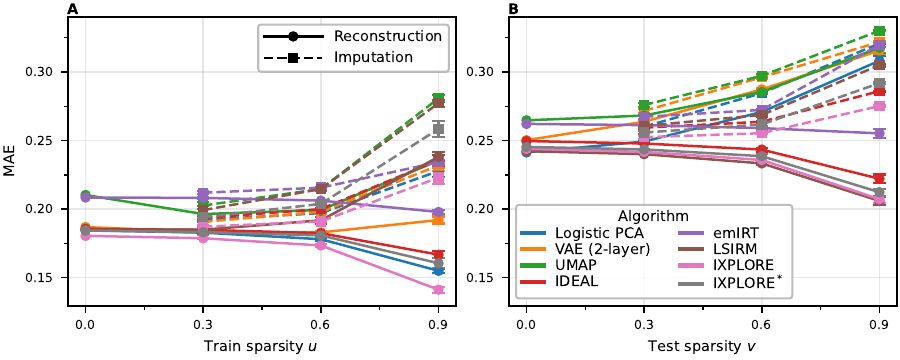}
    \caption{Reconstruction (solid, circles) and imputation (dashed, squares) MAE on \smartvote 2023 as a function of train sparsity $u$ and test sparsity $v$. Error bars show the standard deviation across seeds.}
    \label{fig:baseline_train_test_smartvote_2023}
\end{figure}

\begin{figure}[ht]
    \centering
    \includegraphics{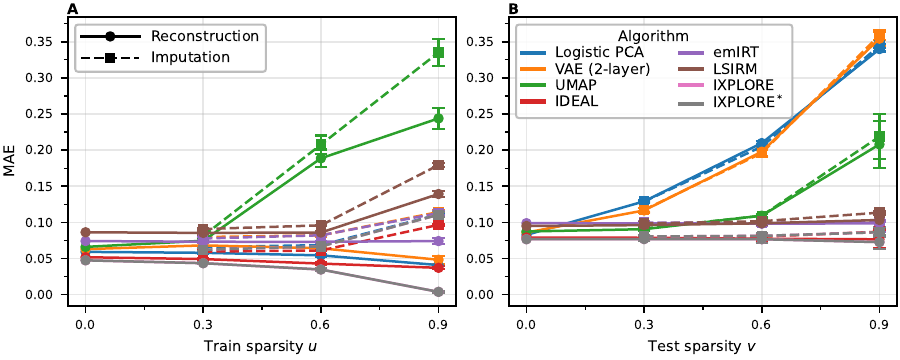}
    \caption{Reconstruction (solid, circles) and imputation (dashed, squares) MAE on \voteview as a function of train sparsity $u$ and test sparsity $v$. Error bars show the standard deviation across seeds.}
    \label{fig:baseline_train_test_voteview}
\end{figure}

\begin{figure}[ht]
    \centering
    \includegraphics{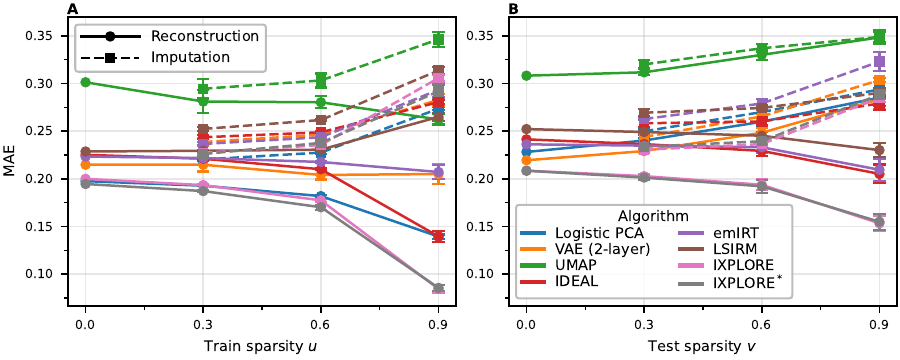}
    \caption{Reconstruction (solid, circles) and imputation (dashed, squares) MAE on \polis as a function of train sparsity $u$ and test sparsity $v$. Error bars show the standard deviation across seeds.}
    \label{fig:baseline_train_test_polis}
\end{figure}

\begin{figure}[ht]
    \centering
    \includegraphics{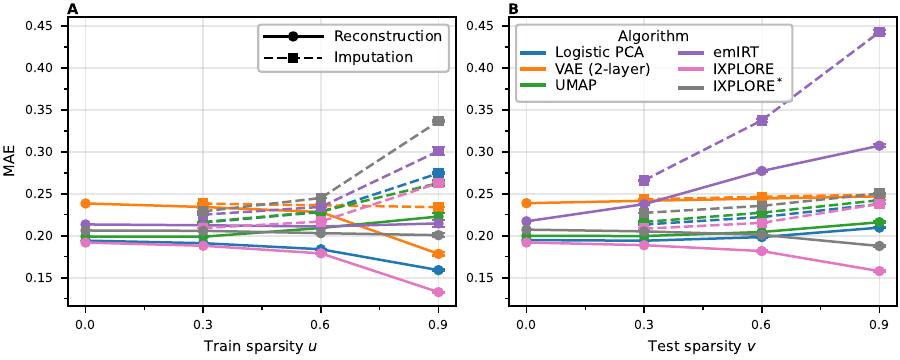}
    \caption{Reconstruction (solid, circles) and imputation (dashed, squares) MAE on EVS as a function of train sparsity $u$ and test sparsity $v$. Error bars show the standard deviation across seeds.}
    \label{fig:baseline_train_test_evs}
\end{figure}
\clearpage
\section{Confidence Weights}
\label{app:weights}

Finally, we ask how users' confidence weights, optionally scaled, affect the imputation error and the latent geometry.
\IXPLORE multiplies each answer's contribution to the user log-likelihood by a per-(user, item) weight $W_{nk}$, and optionally rescales each user's weights so that sparse and dense responders carry the same total weight~\eqref{eq:user-loglik}.
We evaluate three weight schedules derived from the \smartvote 2023 confidence matrix introduced in \cref{sec:datasets}---\emph{uniform} (all 1s), \emph{provided} (the matrix as-is), and \emph{extreme} (the matrix squared)---each with weight scaling disabled and enabled.
The model is trained once with uniform weights at the default configuration and grid resolution $R = 100$; the weights are then applied to the held-out test users at inference time only, across the test-sparsity levels $v$, so the underlying item parameters are held fixed across schedules.

\subsection{Results}
\label{sec:results-weights}

\begin{figure}[ht]
    \centering
    \includegraphics{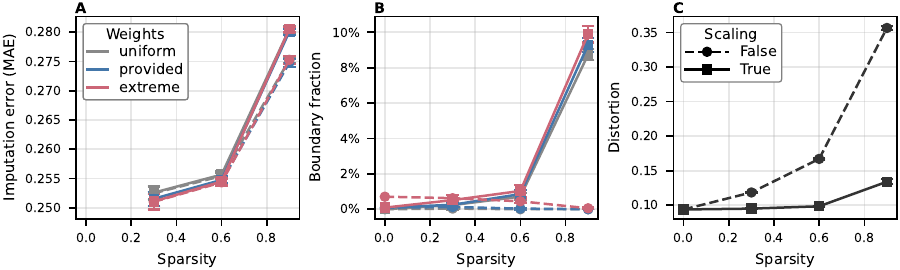}
    \caption{Latent-geometry diagnostics on \smartvote 2023 across test-time sparsity $v$, trained with uniform weights and PCA initialization, $\tau^2 = 0.25$ and $T = 10$; weights are passed at inference time only. The color distinguishes the three weight schedules; line style distinguishes \emph{disabled} and \emph{enabled} weight-scaling. \abc{A} The test imputation error as a function of test-time sparsity. \abc{B} The boundary fraction corresponds to the share of test users within 10\% of the grid boundary. \abc{C} The distortion of the trained latent space is computed by re-embedding synthetic users with uniform weights.}
    \label{fig:weights_geometry}
\end{figure}

\cref{fig:weights_geometry} shows three diagnostics as a function of test-time sparsity.
The weight schedule itself has a negligible effect on the reconstruction and imputation error.
Across \emph{uniform}, \emph{provided}, and \emph{extreme} weights the imputation MAE differs by at most 0.0015 within either setting of weight-scaling.

\begin{figure}[t]
    \centering
    \includegraphics{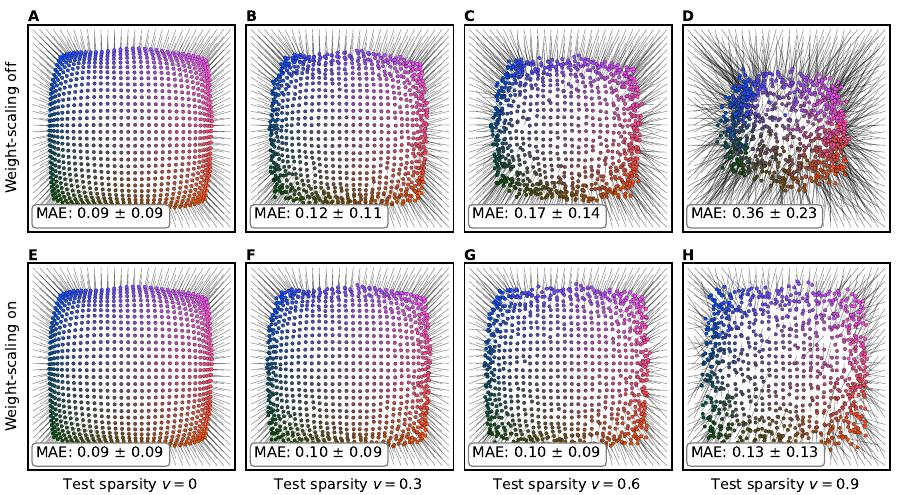}
    \caption{Distortion of the trained latent space on \smartvote 2023 across test-time sparsity $v \in \{0, 0.3, 0.6, 0.9\}$ (columns) with weight scaling disabled (top row) and enabled (bottom row). Synthetic users sampled on a $30 \times 30$ grid have a fraction $1 - v$ of their predicted answers retained and are re-embedded. Each arrow points from a synthetic user's grid position to its re-embedded position; the inset reports the mean and standard deviation of the displacement $\lVert \hat{x} - x \rVert$ over the grid.}
    \label{fig:weights_distortion}
\end{figure}

Weight-scaling, by contrast, changes the latent geometry.
Enabling it pushes sparse users toward the grid boundary, raising the boundary fraction at $v = 0.9$ from 0.0002 to 0.093 (\cref{fig:weights_geometry}B).
At the same time it reduces the distortion of the latent space, with the mean distortion dropping from 0.119 to 0.095 at $v = 0.3$ and from 0.357 to 0.134 at $v = 0.9$ (\cref{fig:weights_geometry}C).
A grid-level visualization of this effect is shown in \cref{fig:weights_distortion}.

\begin{table}[t]
\centering
\caption{Confidence-weights summary on Smartvote 2023. The model is trained once on candidates with uniform weights (PCA initialization, $\tau^2 = 0.25$, $T = 10$ iterations, grid resolution $R = 100$); each row is an inference-time configuration of weight scaling and weight schedule. Cells are averaged across sparsity levels and seeds. The first four metric columns report test reconstruction and imputation error; the last three are latent-geometry diagnostics: the spread as the total variance of the embedding, the boundary fraction (BF), and the distortion (DIS). Best in column \textbf{bold}, runner-up in \textit{italics}.}
\label{tab:weights_smartvote_2023}
\small
\begin{tabular}{llcccccccc}
\toprule
 & & \multicolumn{2}{c}{Test Rec.} & \multicolumn{2}{c}{Test Imp.} & \multicolumn{3}{c}{Latent geometry} \\
\cmidrule(lr){3-4}\cmidrule(lr){5-6}\cmidrule(lr){7-9}
Scaling & Weights & MAE & ACC & MAE & ACC & Spread & BF & DIS \\
\midrule
Disabled & Uniform & 0.229 & 77.5\% & 0.261 & 72.3\% & 0.109 & \textbf{0.0\%} & 0.207 \\
Disabled & Provided & 0.228 & 77.5\% & \textit{0.260} & \textbf{72.4\%} & 0.119 & \textit{0.1\%} & 0.207 \\
Disabled & Extreme & 0.228 & 77.3\% & \textbf{0.260} & \textit{72.4\%} & 0.136 & 0.4\% & 0.207 \\
\midrule
Enabled & Uniform & 0.226 & \textbf{77.8\%} & 0.263 & 72.1\% & 0.156 & 3.0\% & \textbf{0.108} \\
Enabled & Provided & \textbf{0.225} & \textit{77.7\%} & 0.262 & 72.2\% & \textit{0.166} & 3.3\% & 0.108 \\
Enabled & Extreme & \textit{0.225} & 77.6\% & 0.262 & 72.1\% & \textbf{0.181} & 3.6\% & 0.108 \\
\bottomrule
\end{tabular}
\end{table}

\subsection{Discussion}

As a novel extension to ideal point estimation, \IXPLORE allows confidence weights for users to emphasize certain responses.
However, we could not demonstrate the effectiveness of this extension, because the weights provided in the \smartvote 2023 data were not expressive enough: just 11\% of answered items carry a non-default confidence, and the provided values take only two levels (0.5 and 2).
Still, \IXPLORE's \emph{weight-scaling}---the setting that equalizes the total weight of sparse and dense responders---substantially reduces the distortion of sparse users at the cost of a higher boundary fraction.
This could be valuable when displaying users in the political map throughout a survey or deliberation process~\citep{small_polis_2021,bachmann_estimating_2026}, as it would allow a larger spread of users' positions with few responses.

\end{document}